\PassOptionsToPackage{HTML}{xcolor}
\PassOptionsToPackage{numbers,compress}{natbib}
\documentclass{article}

\usepackage{booktabs}
\usepackage{array}
\usepackage{graphicx}
\newcolumntype{R}[1]{>{\raggedright\arraybackslash\hyphenpenalty=10000\exhyphenpenalty=10000}p{#1}}
\usepackage{xcolor}
\usepackage{hyperref}
\hypersetup{colorlinks=true,linkcolor={blue!60!black},citecolor={blue!60!black},urlcolor={blue!60!black},anchorcolor=black,breaklinks=true,hypertexnames=false}
\usepackage{url}
\usepackage{xurl}
\usepackage{amsmath}
\usepackage{amssymb}
\usepackage{enumitem}
\usepackage{placeins}
\usepackage{float}
\usepackage{needspace}
\usepackage{wrapfig}

\newcommand{\tblbest}[1]{#1}
\newcommand{\ci}[2]{#1$\,\pm\,$#2}
\newcommand{\lift}[2]{#1\,(\,#2\,)}

\newcommand{\frameworkname}{\textbf{StagedWorkspace}}

\newcommand{\systemshort}{\textbf{\textsc{SW-Agent}}}
\newcommand{\tablefontsize}{\footnotesize}

\newcommand{\resulttablesetup}{%
  \tablefontsize
  \setlength{\tabcolsep}{3.5pt}%
  \renewcommand{\arraystretch}{1.02}%
}
\newcommand{\appendixtablesetup}{%
  \footnotesize
  \setlength{\tabcolsep}{2.0pt}%
  \renewcommand{\arraystretch}{0.90}%
}
\usepackage{tabularx}
\usepackage{subcaption}
\newcommand{\apexldb}{\textsc{APEX-Agents} public leaderboard}
\ifdefined\artifactstatepreprint
  \PassOptionsToPackage{preprint}{neurips_2026}
  \usepackage{float} % [H] on generated narrative floats (inline with prose)
  \let\FloatBarrier\relax
\fi
\IfFileExists{neurips_2026.sty}{%
  \usepackage[preprint]{neurips_2026}%
}{%
  \usepackage[numbers, compress]{natbib}%
}
\ifdefined\artifactstatepreprint
\fi

\usepackage[utf8]{inputenc} % allow utf-8 input
\usepackage[T1]{fontenc}    % use 8-bit T1 fonts
\usepackage{amsfonts}       % blackboard math symbols
\usepackage{nicefrac}       % compact symbols for 1/2, etc.
\usepackage[expansion=false]{microtype} % microtypography; avoids TeX Live 2025 \showhyphens warning
\usepackage{fvextra} % breakable verbatim (grader prompts in appendix)
\fvset{breaklines=true,breakanywhere=false,breaksymbol=,breaksymbolleft=,breaksymbolright=,fontsize=\footnotesize}

\providecommand{\answerYes}{\textbf{Yes}}
\providecommand{\answerNo}{\textbf{No}}
\providecommand{\answerNA}{\textbf{N/A}}

\newcommand{\missingfigure}[1]{%
  \fbox{\parbox[c][0.22\textheight][c]{0.92\linewidth}{\centering Missing figure file:\\[0.5em]\texttt{\detokenize{#1}}}}%
}
\newcommand{\includegraphicsfallback}[3][]{%
  \IfFileExists{#2}{\includegraphics[#1]{#2}}{%
    \IfFileExists{#3}{\includegraphics[#1]{#3}}{\missingfigure{#2}}%
  }%
}

\ifdefined\artifactstatepreprint
  \title{\frameworkname{}: Managing Workspace State for Knowledge-Work Agents\\[-0.2em]{\large Preprint}}
\else
  \title{\frameworkname{}: A Versioned Workspace for Knowledge-Work Agents}
\fi

\author{
  Yining Hua\textsuperscript{1,2} \quad
  Hongbin Na\textsuperscript{3} \quad
  Yifan Zhou\textsuperscript{2,4} \quad
  Akshay Kalose\textsuperscript{2,5} \\
  \textbf{Cyrus Ayubcha}\textsuperscript{1} \quad
  \textbf{Levi Lian}\textsuperscript{2,5}\thanks{Correspondence to Levi Lian: \texttt{levi@raycaster.ai}} \\[2pt]
  \textsuperscript{1}Harvard University \quad
  \textsuperscript{2}Raycaster AI \quad
  \textsuperscript{3}University of Technology Sydney \\
  \textsuperscript{4}University of Washington \quad
  \textsuperscript{5}Stanford University
}

\begin{document}

\maketitle

\begin{abstract}
AI agents increasingly perform knowledge work (i.e., produce and modify persistent digital artifacts such as code repositories, documents, spreadsheets, slides, reports), yet the parsed views they search, the native files they edit, the changes they review, and the artifacts they submit can refer to different versions of the same work product. We formulate this as a workspace-state contract: every view should be explicitly tied to a version of the evolving workspace state. Coding agents partly address this need through repository contracts for search, diffs, and tests, whereas an analogous contract is less explicit for PDFs, spreadsheets, slides, notebooks, and mixed-format project folders. We propose \frameworkname{}, a versioned workspace for knowledge-work agents. The workspace binds parsed records and review diffs to content hashes of the native files as they change. In fixed-harness ablations on \textsc{OfficeQA Pro} and \textsc{APEX-Agents}, dual parsed/native access has the highest point estimate for every tested model; relative to the more limiting single view, it improves OfficeQA Pass@1 by 8.3--12.1 points and APEX mean rubric score by 4.7--9.2 points. \systemshort{} scores 63.9\% with Gemini~3.1 Pro on OfficeQA and 42.1 with GPT-5.4 Nano on APEX, compared with published same-model scores of 29.3\% and 25.5, respectively. A paired review-axis ablation on 57 file-editing tasks further finds higher observed scores when diffs are visible. These results identify workspace state as an experimental variable in knowledge-work agents and motivate benchmarks that score evidence, staged edits, and submitted artifacts as explicit state transitions.
\end{abstract}

\section{Introduction}
\label{sec:introduction}

Recent large language model (LLM) agent research has moved from isolated text-only tasks toward work grounded in digital artifacts: coding agents repair repositories \citep{jimenez2023swebench,yang2024sweagent}, document and office agents answer questions or edit files \citep{suri2024doceditv2,wang2024officebench}, and web, computer-use, and workplace agents operate across browsers, desktops, and enterprise tools \citep{xie2024osworld,xu2024theagentcompany}. These settings differ, but they share a target: knowledge work. Classic accounts define knowledge work as producing or transforming information through interpretation, judgment, and revision \citep{drucker1999knowledgeworker,davenport2005thinking}. Following \citet{hua2026designreport}, we treat coding, document, office, web, and workplace agents as branches of this broader class, in which an agent is ultimately judged by the work product it leaves behind.

Work-product evaluations judge persistent outputs such as answers, edits, or deliverables against criteria applied by a grader or reviewer \citep{hua2026designreport,officeqa2026,apexagents2026}. An agent can otherwise retrieve a column definition from one workbook version, edit another version in its sandbox, and submit a deliverable inconsistent with the evidence it used. This motivates a workspace-state contract: parsed search, native operations, review, and submission must each identify the workspace version they expose or modify. Coding agents partly instantiate such a contract through a repository checkout that keeps search, edits, and tests in a shared state \citep{jimenez2023swebench,yang2024sweagent}. Non-coding workspace interfaces do not always expose an equivalent
contract, even though their tasks still require source reading, mixed-format project work, and materialized deliverables \citep{officeqa2026,apexagents2026,tang2026workspacebench}. Without that contract, parsed search, native edits, and the submitted file can drift to different versions of the same work product.

The need for such a contract arises from three recurring interface trade-offs. First, artifact-only systems preserve complete native artifacts but make search difficult, forcing agents to page through large files and carry irrelevant context. Second, parsed-only systems support search but can lose layout hierarchy, formulas, and visual evidence. Third, an unversioned mutable workspace lets an agent overwrite, move, or delete files without a durable diff for model or human review. Together, these trade-offs motivate two separable interface requirements: complementary parsed and native views tied to explicit file versions, and a reviewable record of changes to the submitted state.

\begin{wrapfigure}{r}{0.50\columnwidth}
  \vspace{-8pt}
  \centering
  \includegraphics[width=\linewidth]
    {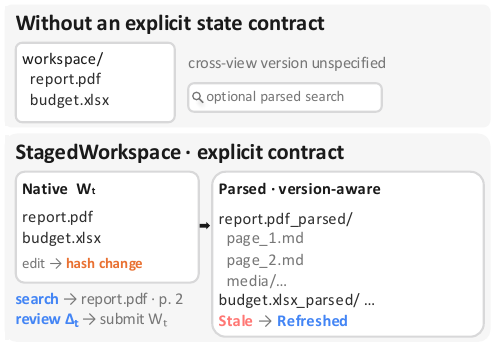}
  \caption{An explicit workspace-state contract ties native artifacts,
  parsed views, review state, and submission to the evolving workspace.}
  \label{fig:workspace-contract}
  \vspace{-10pt}
\end{wrapfigure}

In response, we propose \frameworkname{}, a workspace layer for knowledge-work agents built around an explicit workspace-state contract.  As illustrated in Figure~\ref{fig:workspace-contract}, \frameworkname{} provides complementary parsed and native artifact views and ties parsed records, staged changes, review state, and submitted artifacts to the same evolving workspace state. We make three core contributions: (1) we define a workspace-state contract for mixed-format knowledge work; (2) we instantiate it in \systemshort{} through version-aware artifact views, source-hash-based cache invalidation and refresh, and journaled review diffs; and (3) we evaluate \systemshort{} on \textsc{APEX-Agents} \citep{apexagents2026} and \textsc{OfficeQA Pro} \citep{officeqa2026} using matched ablations that isolate the roles of artifact-view availability and review visibility.

\section{Related Work}
\label{sec:relwork}

\paragraph{Knowledge-Work Agents and Work-Product Evaluation.}
\citet{hua2026designreport} frame knowledge-work benchmarks around a work activity, a tested setting, and a scored work product. The gap is not a lack of benchmarks. WorkArena and WorkArena++ emphasize enterprise web workflows \citep{drouin2024workarena,boisvert2024workarenaplusplus}; TheAgentCompany and OSWorld stress workplace or desktop control \citep{xu2024theagentcompany,xie2024osworld}; OfficeBench and OdysseyBench cover office automation across documents, spreadsheets, and longer workflows \citep{wang2024officebench,odysseybench2025}; \textsc{APEX-Agents} and Workspace-Bench add larger project folders and materialized deliverables \citep{apexagents2026,tang2026workspacebench}; and FileGram studies behavioral traces over file systems \citep{liu2026filegram}. These settings exercise pieces of knowledge work, but they usually score a task, environment, trace, or final product without making the shared artifact state itself the experimental object. \frameworkname{} targets that missing contract.

\paragraph{Agent Infrastructure and Retrieval.}
ReAct interleaves model reasoning with external actions, allowing a model to choose its next step based on intermediate observations \citep{yao2023react}. Retrieval-augmented generation conditions its output on retrieved passages \citep{Lewis2020RAG}, and dense passage retrieval supplies the neural lookup that finds those passages in open-domain corpora \citep{karpukhin-etal-2020-dense}. These methods govern how an agent obtains context, but do not specify the version relation among mutable artifacts: parsed search results, native files, staged edits, and the final submission are never required to refer to one workspace version. Retrieval can then return evidence from a snapshot that the agent has already changed. \frameworkname{} makes that relation explicit.

\paragraph{Patch-Based Software-Engineering Agents.}
Software-engineering agents come closest to a shared workspace-state contract. In SWE-bench, a repository snapshot defines source state, and a test suite gives partial feedback on a proposed patch \citep{jimenez2023swebench}, and SWE-agent exposes patch-oriented tools over that same repository interface \citep{yang2024sweagent}. What transfers from this design is the convention: the agent, evaluator, and reviewer agree on how to represent the current state, proposed changes, and partial verification. It assumes line-addressable text and exact search, both of which are available for source code. Other knowledge-work projects break that assumption: they mix binary office files, notebooks, and layout-heavy PDFs that cannot be diffed line by line. For these formats, the parsed, searchable view must stay aligned with the native copy the agent edits and submits, a coupling that the repository interface does not provide.

\paragraph{Document Editing and Versioned Artifacts.}
The mechanisms that would supply such a contract appear in document AI and versioned authoring, but each covers only part of it. Document AI studies layout-aware question answering over scanned documents \citep{mathew2021docvqa} and long research papers \citep{dasigi2021qasper}, which addresses reading rather than editing. A separate line studies localized document editing \citep{mathur2023docedit}, layout-grounded edit commands \citep{suri2024doceditv2}, and document-level model editing \citep{zeng2025docmedit}. Verifiable-editing systems make model edits inspectable through edit operations \citep{goosvc2025}, warnings and verification steps \citep{laban2024inksync}, or process traces \citep{draftmarks2025}, and anchored feedback ties suggestions to specific text spans to support localized revision \citep{anchoredai}. Versioned authoring tools record replayable manuscript histories \citep{himmelstein2019manubot}, AI-assisted manuscript updates \citep{pividori2024manubotgpt}, and version-controlled transcription editing \citep{zhang2017gitdox}, while research-data systems version datasets and track provenance for review \citep{halchenko2021datalad,falster2019datastorr}. These systems give a human author edit inspection and version history. None of them keep a parsed, machine-readable cache synchronized in real time with the native files an agent searches, edits, and submits, which is the coupling \frameworkname{} contributes.

\section{Methods}
\label{sec:method}
\label{sec:substrate}

\subsection{\frameworkname{}: A Version-controlled Workspace}
\label{sec:artifactstate-contract}
\label{sec:workspace-abstraction}

Knowledge-work agents often require complementary representations of the same artifact. Parsed representations support search and evidence localization, while native artifacts preserve layout, formulas, executability, and the object that is ultimately edited or submitted. \frameworkname{} therefore keeps different workspace file views tied to the same evolving file state.

\begin{figure}[!ht]
\centering
\includegraphicsfallback[width=\linewidth,height=0.8\textheight,keepaspectratio]{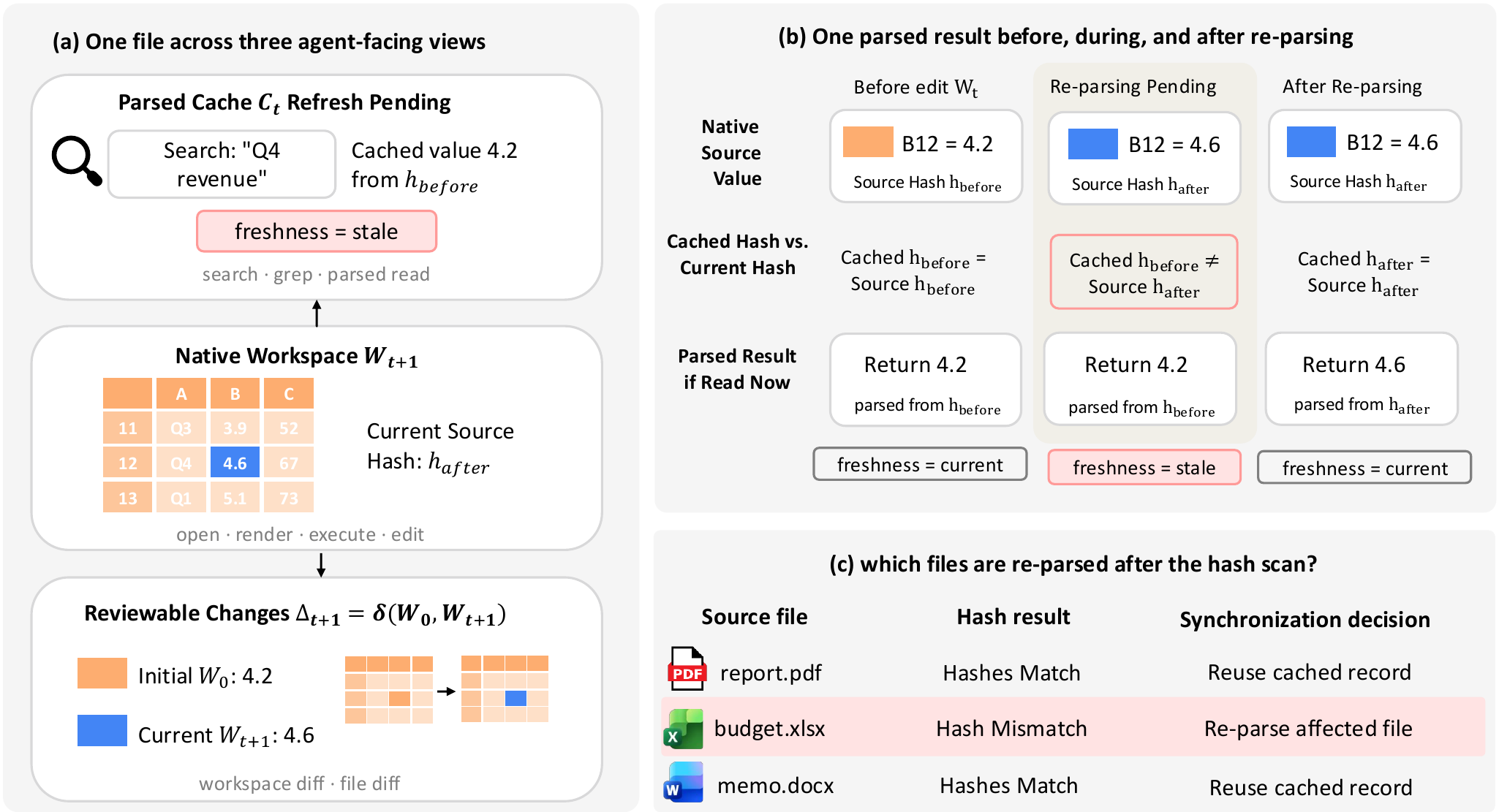}{Figs/ArtifactState_Workspace.png}

\caption{Workspace synchronization across three agent-facing views. (a) After editing budget.xlsx at Sheet1!B12 and scanning content hashes, the native workspace contains $W_{t+1}$, the review view compares $W_0$ with $W_{t+1}$, and a parsed read may still return the prior value with freshness = stale. (b) The parsed result remains stale while $h_{\mathrm{before}}\ne h_{\mathrm{after}}$ and becomes current after asynchronous re-parsing. (c) Matching hashes reuse cached records; in this example, only the mismatched budget.xlsx is re-parsed.}
\label{fig:artifactstate-architecture}
\end{figure}

Figure~\ref{fig:artifactstate-architecture} gives the overall view of \frameworkname{}. The task corpus, policy prompt, tool-call budget, and grading rubric are held fixed across conditions, so the read-axis ablations change only the artifact interface.

At synchronization point $t$, taken after a tool batch, \frameworkname{} represents a task with three agent-facing views:
\[
\begin{aligned}
  W_t &= \text{current native workspace files},\\
  C_t &= \text{parsed records tagged by source path and hash},\\
   \Delta_t &= \delta(W_0, W_t).
\end{aligned}
\]

The native workspace $W_t$ is the authoritative state used for execution and submission. Each record in $C_t$ identifies the source path and content hash of the file version from which it was parsed. A record is current when its source hash matches the corresponding file in $W_t$; otherwise it is explicitly marked stale until refreshed. Parsed tools read $C_t$, native operations read or edit $W_t$, and review tools expose $\Delta_t$, the changes between the starting workspace $W_0$ and the current workspace $W_t$.

The parsed cache and review diff are derived views, not separate document copies. Internally, the workspace also keeps an accepted artifact tree and a staged-change journal. The journal adds operation-level state on top of the workspace snapshots, allowing \frameworkname{} to detect stale edits, expose pending changes for review, and promote or roll back accepted changes.

Appendix~\ref{app:agent-artifact-layer} specifies the parsed, visual, and native tool surfaces, and Appendix~\ref{app:implementation-details} gives the lower-level journal, promotion, and conflict rules. 
After each batch of mutations, \frameworkname{} synchronizes the sandbox back to the workspace state. A hash scan identifies changed native files, updates $W_t$, and refreshes the affected parsed records in $C_t$:
\[
\begin{aligned}
  W_t &\rightarrow W_{t+1},\\
C_{t+1} &= \operatorname{sync}(C_t,W_{t+1}),\\
\Delta_{t+1} &= \delta(W_0, W_{t+1}).
\end{aligned}
\]

Here, $\operatorname{sync}$ immediately marks hash-mismatched records stale and replaces them when asynchronous parsing completes. Thus, the agent can search parsed files, open the corresponding originals, and inspect edits against the file state that will be submitted.

\subsection{Agent Turn Loop}
\label{sec:eval-agent-turn}

The state model above defines what must stay synchronized; the agent turn loop defines when synchronization occurs. Each benchmark item starts by materializing task files into $W_t$ and recording path-level content hashes. The agent then acts in a ReAct-style tool loop until it submits or reaches the tool budget.

At the beginning of a turn, the sandbox is hydrated from the current $W_t$. After each mutating tool batch, the harness hash-scans the sandbox and advances the workspace to point $t{+}1$ through the update defined in Section~\ref{sec:artifactstate-contract}. Changed hashes invalidate only the affected parsed records before the next parsed observation.

This protocol gives the agent a version-aware read-after-write contract: after an edit, native operations resolve against the updated workspace, while parsed results either match that version or are labeled stale until refreshed. Before the final handoff, the submitted answer or deliverable is produced from the staged workspace state. When diff tools are visible, the agent can inspect $\Delta_t$ before submission; journals, diffs, and traces are retained for replay and post-hoc trajectory analysis.

The ablations perturb this operational contract through three mechanisms.

\subsection{Synchronized dual artifact access}
\label{sec:artifactstate-mechanisms}
\label{sec:execution-flow}
\label{sec:protocol}
\label{sec:setup-harness}

Dual artifact access is the primary read interface of StagedWorkspace: parsed views support evidence localization, while native artifacts support faithful inspection, execution, and editing. The loop above combines the three workspace mechanisms tested or held fixed in the ablations:

\emph{Dual artifact views} expose two synchronized read paths over the same $W_t$. Parsed search and parsed reads localize evidence across large document collections. Native reads preserve layout, formulas, and executable office files. Because both paths derive from the same workspace state, the agent can search a parsed cache and then open or execute the corresponding native file without crossing artifact versions.

\emph{Hash-keyed synchronization} keeps the parsed cache as derived state. Each parsed record is keyed to the source path and content hash. When sandbox synchronization observes a changed hash, \frameworkname{} marks only the affected parsed records stale and queues them for refresh. Unchanged files reuse their prior parsed records. This rule prevents vector search, grep, and parsed reads from presenting evidence from an earlier file version as current.

\emph{Journaled review} records staged edits in the same workspace state that drives submission. The review surface $\Delta_t$ is computed from the starting and current workspace files, so \texttt{workspace\_diff} and \texttt{workspace\_file\_diff} show the edits that would be handed off. In our implementation, this surface is format-specific: text files use line diffs, spreadsheets expose row- and cell-level changes, slide decks expose slide-level changes, and unsupported binaries fall back to before/after file previews. Appendix~\ref{app:substrate-formal} describes the format-dependent diff surface.

\FloatBarrier
\subsection{Experimental Design}
\label{sec:setup}
\label{sec:benchmarks}

We use one full-system benchmark comparison and two mechanism ablations. The comparison runs \frameworkname{}, implemented in \systemshort{}, under the original benchmark graders. The read-axis ablation tests synchronized dual views against artifact-only and parsed-only access while holding the model, prompt, parser, retriever, grader, file tracker, and tool budget fixed. The review-axis ablation changes only whether tracked diffs are visible before submission. We report latency and agent-run cost as diagnostics, not as primary outcomes.

\subsubsection{Complete-System Benchmark Study}
\label{sec:complete-system-study}
\label{sec:setup-benchmarks}

The full-system comparison evaluates \systemshort{} with the complete \frameworkname{} workspace, including synchronized parsed and native views. It measures end-to-end task performance under the benchmark protocols. The mechanism claims come from the ablations in Section~\ref{sec:setup-ablation}.

We evaluate on two public knowledge-work agent benchmarks with published graders and task pools, without retrofitting corpora or rubrics. The two benchmarks span complementary regimes: \textsc{OfficeQA Pro} tests read-heavy corpus QA, with full-library retrieval and numerical grounding over a fixed document collection; \textsc{APEX-Agents} tests deliverable-heavy professional work, with cross-format project folders and multi-criterion rubric grading on materialized outputs.

\begin{itemize}
    \item \textbf{\textsc{OfficeQA Pro}.}
    \textsc{OfficeQA Pro} is a grounded numerical QA benchmark over the U.S.\ Treasury \emph{Treasury Bulletin}~\citep{officeqa2026}: $104$ document-only questions and $29$ questions that require live web evidence.
    Each evaluation loads a shared workspace of ${\approx}697$ bulletin PDFs (1939--2025); agents must locate one or a few sources per question.
    The primary outcome is exact-match Pass@1 over the benchmark question set.
    The internet subset uses web tools, while the remaining questions are document-only.
    Efficiency diagnostics add the benchmark's released text-export baseline as context.

    \item \textbf{\textsc{APEX-Agents}.}
    \textsc{APEX-Agents} measures cross-application professional work across 33 worlds (Management Consulting, Investment Banking, Law) with 480 rubric-graded tasks; each task folder averages ${\approx}166$ mixed-format files~\citep{apexagents2026}. We exclude the 28 tasks in Investment Banking Worlds 244 and 246 because they depend on external APIs (EDGAR) that were not configured.
    Each task has 1--10 pass/fail criteria (4.06 on average), evaluated on the materialized deliverable.
    The primary outcomes are task pass rate and mean rubric score.
\end{itemize}

Published benchmark rows are included only as contextual comparisons because they use different harnesses, corpora, or attempt budgets. The paired ablation studies below provide the controlled evidence for the workspace mechanisms.

Both benchmarks use the same \systemshort{} harness for our runs: ReAct-style interleaved reasoning and tool calls until submission or a 250-tool-call budget, which remains unchanged across artifact-view ablations.
On \textsc{APEX-Agents}, that loop matches the public protocol; published \textsc{OfficeQA Pro} rows use the benchmark's custom agent~\citep{officeqa2026}.
We quote those published rows without rerunning the benchmark agent, but run our ablations on the same question set with matched models and the same exact-match grader.
Uploads or journaled edits enqueue hash-keyed re-parsing and re-indexing, so the parsed view stays aligned with the sandbox files. Appendix~\ref{app:agent-runtime} gives the tool-routing details and the released text-export baseline.

\subsubsection{Mechanism Ablation Studies}
\label{sec:setup-ablation}
\label{sec:setup-protocol}
\label{sec:setup-diff-ablation}

Mechanism studies isolate two workspace components. The read-axis ablation maps the formal workspace views to experimental arms: dual exposes both the native working tree $W_t$ and the hash-keyed parsed cache $C_t$; artifact-only exposes only $W_t$; and parsed-only exposes only $C_t$. Records in $C_t$ are keyed to their source paths and content hashes; a record whose source hash no longer matches the corresponding file in $W_t$ is marked stale until refreshed. Dual and parsed-only share the same parsed index; artifact-only disables access to the parsed index. File tracking, prompts, parser, retriever, grader, and tool budget are fixed across these arms. The review-axis ablation holds dual views fixed and changes only whether the agent can inspect tracked file changes before submission.

\begin{table}[!htbp]
  \caption{Read-axis artifact-view arms. Original artifacts are the hydrated workspace source files $W_t$; the parsed index is the hash-keyed searchable cache $C_t$, with each record labeled current or stale according to whether its source hash matches $W_t$.}
  \label{tab:ablation-design}
  \centering
  \resulttablesetup
  \begin{tabularx}{\linewidth}{@{}lcccX@{}}
    \toprule
    \textbf{Arm} & \textbf{Originals} & \textbf{Parsed search} & \textbf{Hash sync} & \textbf{Mechanism isolated} \\
    \midrule
    Dual &
    Yes ($W_t$) &
    Yes ($C_t$) &
    Yes &
    Synchronized native and parsed access. \\
    \addlinespace
    Artifact-only &
    Yes ($W_t$) &
    No &
    N/A &
    Native access without indexed parsed search. \\
    \addlinespace
    Parsed-only &
    No &
    Yes ($C_t$) &
    Yes &
    Parsed search without native inspection or execution. \\
    \bottomrule
  \end{tabularx}
\end{table}

File tracking is enabled in all three read-axis arms; only the read views change. The review-axis ablation is separate: both conditions use dual views and identical file tracking, but only the diff-visible arm exposes \texttt{workspace\_diff} and \texttt{workspace\_file\_diff} before submission. OfficeQA is excluded from the review-axis ablation because it scores final text answers rather than edited files.

Main-text read-axis tables include only models with completed three-arm runs, because a paired contrast needs dual, artifact-only, and parsed-only for the same model; partial sweeps are dropped by this completeness rule, not by their scores.

\subsubsection{Metrics and Uncertainty}
\label{sec:setup-metrics}

We use the outcome definitions supplied by each benchmark. OfficeQA reports exact-match accuracy over its question set~\citep{officeqa2026}; each run has one graded attempt per question, so this accuracy is also run-level Pass@1. APEX reports task pass rates and rubric scores over professional-work tasks~\citep{apexagents2026}; a task passes only when all rubric criteria pass on the materialized deliverable, and mean rubric score is a task-uniform average under the benchmark rubric.

Attempt budgets differ by run role. Models in the read- and review-axis ablations, the GPT-5.4 family and Gemini~3~Flash, use three independent attempts per task, which keeps the paired contrasts on a uniform budget for statistical testing. Context-only dual runs that appear only in the published comparisons, Gemini~3.1~Pro and Kimi~K2.6, use a single attempt. We report cost and latency only as diagnostic quantities.

All uncertainty intervals are nonparametric bootstrap intervals computed over the benchmark unit: questions for OfficeQA and tasks for APEX. We draw 10{,}000 resamples with seed~20260515 and report point estimates with the half-width of the central 95\% interval, computed from the 2.5/97.5 percentile endpoints and rounded to one decimal point. These half-widths are uncertainty intervals, not sample standard deviations. For paired ablations, each bootstrap sample resamples matched items and recomputes the within-item contrast. This preserves the covariance between arms and is the primary uncertainty estimate for read-axis effects.

\section{Results}
\label{sec:results}
\label{sec:experiments}

\subsection{Complete-System Benchmark Results}
\label{sec:complete-system-results}

We first place the full \systemshort{} stack in the original benchmark settings. Table~\ref{tab:complete-system-results} compares \systemshort{} with public leaderboard results on the same model where available, plus the strongest available reference row where the benchmark reports one. These comparisons locate the system against published agents; they do not identify causality. The fixed-harness ablations in Table~\ref{tab:mechanism-ablation-results} provide the one-to-one controls over workspace mechanisms.

On \textsc{OfficeQA Pro}, the main comparison is against the best published \emph{Full} row for the same model. Here \emph{Full} means the full Treasury corpus is available, rather than the Oracle setting that receives gold source documents; \emph{dual views} means \systemshort{} exposes both native files and parsed search over that full corpus. \systemshort{} improves over the best published Full row by 8.3 Pass@1 points for GPT-5.4 (64.7 vs.\ 56.4), 34.6 points for Gemini~3.1 Pro (63.9 vs.\ 29.3), and 24.7 points for Gemini~3 Flash (57.8 vs.\ 33.1). The Oracle rows remain useful references, but they are an easier source-injected setting without the complexity that comes with search and retrieval, and should be interpreted as the ceiling.

On \textsc{APEX-Agents}, \systemshort{} is compared with the matched published leaderboard row for each model. The largest gains are for GPT-5.4 Nano (+8.1 Pass@1 and +16.6 mean-score points) and Gemini~3 Flash (+6.9 Pass@1 and +8.3 mean-score points). The gains exist for frontier models too: GPT-5.4 improves by +1.7 Pass@1 and +0.9 mean-score points, and Gemini~3.1 Pro improves by +2.6 and +2.3 points. The GPT-5.4 \systemshort{} row also exceeds the Opus 4.7 reference row reported by the benchmark (37.7 vs.\ 33.9 Pass@1; 53.6 vs.\ 50.6 mean score).

\begin{table}[!htbp]
  \caption{Complete-system comparisons. In Panel~(a), ``Full corpus + dual views'' means \systemshort{} runs on the full OfficeQA corpus with native files and parsed search; Oracle rows receive gold sources. Panel~(b) includes matched APEX rows and an Opus reference. N/A = not reported; -- = not applicable.}
  \label{tab:complete-system-results}
  \centering
  \begin{subtable}[t]{\linewidth}
  \caption{\textsc{OfficeQA Pro}: quoted public rows vs. dual \systemshort{} on matched models. Comparisons are non-paired.}
  \label{tab:officeqa-reported-comparison}
  \centering
  \resulttablesetup
  \setlength{\tabcolsep}{2.4pt}
  \begin{tabular}{@{}llrrrr@{}}
    \toprule
    \textbf{System} & \textbf{Configuration} & \textbf{Acc.} & \textbf{Latency} & \textbf{Tools} & \textbf{Cost} \\
    \midrule
    \multicolumn{6}{@{}l}{\textit{GPT-5.4}} \\
    OfficeQA & Full corpus + artifact-only & 36.1 & 13.1m & 57.0 & \$1.79 \\
    OfficeQA & Full corpus + parsed (Databricks) & 56.4 & 3.6m & 34.5 & \$1.26 \\
    OfficeQA & Oracle sources + artifact-only & 54.9 & 4.4m & 31.6 & \$0.44 \\
    OfficeQA & Oracle sources + parsed (Databricks) & 65.4 & 2.2m & 20.7 & \$0.33 \\
    \systemshort{} & Full corpus + dual views & \ci{64.7}{8.3} & 6.1m & 30.4 & \$0.84 \\
    \midrule
    \multicolumn{6}{@{}l}{\textit{Gemini 3.1 Pro}} \\
    OfficeQA & Full corpus + artifact-only & 18.1 & 26.4m & 75.2 & \$6.21 \\
    OfficeQA & Full corpus + parsed (Databricks) & 29.3 & 2.9m & 12.8 & \$1.61 \\
    OfficeQA & Oracle sources + artifact-only & 39.1 & 4.2m & 28.4 & \$0.76 \\
    OfficeQA & Oracle sources + parsed (Databricks) & 46.6 & 2.6m & 11.4 & \$0.23 \\
    \systemshort{} & Full corpus + dual views & \ci{63.9}{7.9} & 10.3m & 25.9 & \$0.64 \\
    \midrule
    \multicolumn{6}{@{}l}{\textit{Gemini 3 Flash}} \\
    OfficeQA & Full corpus + artifact-only & N/A & N/A & N/A & N/A \\
    OfficeQA & Full corpus + parsed (Databricks) & 33.1 & 3.0m & 40.6 & \$0.79 \\
    \systemshort{} & Full corpus + dual views & \ci{57.8}{8.6} & 15.3m & 39.8 & \$0.48 \\
    \bottomrule
  \end{tabular}
\end{subtable}

  \vspace{0.75em}
  \begin{subtable}[t]{\linewidth}
  \caption{\textsc{APEX-Agents}: dual \systemshort{} vs. published rows. Comparisons are non-paired.}
  \label{tab:apex-reported-comparison}
  \centering
  \resulttablesetup
  \setlength{\tabcolsep}{2.4pt}
  \begin{tabular}{@{}llrrrr@{}}
    \toprule
    \textbf{System} & \textbf{Match} & \textbf{Pass@1}$^{\dag}$ & \textbf{Mean} & \textbf{$\Delta$ Pass@1} & \textbf{$\Delta$ mean} \\
    \midrule
    Kimi K2.6 & vendor row & 27.9 & \multicolumn{1}{r}{N/A} & -- & -- \\
    \systemshort{} Kimi K2.6 & same model & \ci{31.2}{4.7} & \multicolumn{1}{r}{N/A} & +3.3 & \multicolumn{1}{r}{N/A} \\
    \midrule
    APEX GPT-5.4 Nano (xHigh) & same model & \ci{16.9}{2.6} & \ci{25.5}{2.6} & -- & -- \\
\systemshort{} GPT-5.4 Nano & same model & \ci{25.0}{3.4} & \ci{42.1}{4.1} & +8.1 & +16.6 \\
    \midrule
    APEX Gemini 3 Flash (High) & same model & \ci{24.0}{3.3} & \ci{39.5}{3.3} & -- & -- \\
    \systemshort{} Gemini 3 Flash & same model & \ci{30.9}{2.6} & \ci{47.8}{2.4} & +6.9 & +8.3 \\
    \midrule
    APEX GPT-5.4 Mini (xHigh) & same model & \ci{24.6}{3.2} & \ci{37.5}{3.1} & -- & -- \\
\systemshort{} GPT-5.4 Mini & same model & \ci{28.2}{3.5} & \ci{44.3}{4.4} & +3.6 & +6.8 \\
    \midrule
    APEX Gemini 3.1 Pro (High) & same model & \ci{33.5}{3.6} & \ci{48.2}{3.4} & -- & -- \\
    \systemshort{} Gemini 3.1 Pro & same model & \ci{36.1}{3.8} & \ci{50.5}{3.3} & +2.6 & +2.3 \\
    \midrule
    APEX GPT-5.4 (xHigh) & same model & \ci{36.0}{3.8} & \ci{52.7}{3.4} & -- & -- \\
    \systemshort{} GPT-5.4 & same model & \ci{37.7}{4.0} & \ci{53.6}{3.5} & +1.7 & +0.9 \\
    \midrule
    APEX Opus 4.7 & reference & \ci{33.9}{3.8} & \ci{50.6}{3.5} & -- & -- \\
    \bottomrule
  \end{tabular}
\end{subtable}

\end{table}

The two benchmarks expose complementary failure modes for agentic knowledge work. \textsc{OfficeQA Pro} exposes the challenges of search, source localization and table-grounded reading over a large PDF collection, while \textsc{APEX-Agents} stresses the combination of evidence retrieval in a massive enterprise corpus and implementing professionally valuable work. The mechanism ablations below test whether those gains come from synchronized artifact views and visible review state rather than from the benchmark setting alone.

\subsection{Mechanism Ablations}
\label{sec:mechanism-ablation-results}
\label{sec:ablation}
\label{sec:diff-ablation}

The ablations test two workspace mechanisms from Section~\ref{sec:setup-ablation}: synchronized read views and visible review diffs. We report dual vs.\ artifact-only vs.\ parsed-only on both benchmarks, and diffs visible vs.\ hidden on \textsc{APEX} only, since OfficeQA tasks do not involve file editing.

\begin{table}[!htbp]
  \caption{Mechanism ablations. Panel~(a) reports the \textsc{OfficeQA Pro} read-axis ablation using Pass@1. Panel~(b) reports the \textsc{APEX-Agents} read-axis ablation using mean rubric score. Panel~(c) reports the \textsc{APEX-Agents} review-axis ablation on paired file-editing tasks. In Panels~(a) and~(b), lift columns show signed paired lift from dual with the bootstrap half-width in parentheses.}
  \label{tab:mechanism-ablation-results}
  \centering

  \begin{subtable}{\linewidth}
    \centering
    \caption{\textsc{OfficeQA Pro}: read-axis ablation. Pass@1 is shown as point $\pm$ question-bootstrap 95\% half-width.}
    \label{tab:read-axis-summary}
    \label{tab:officeqa-dual-artifact-ablation}
    \resulttablesetup
    \begin{tabular}{lrrrrr}
      \toprule
      \textbf{Model} & \textbf{Dual} & \textbf{Artifact-only} & \textbf{Parsed-only} & \textbf{Lift vs art.} & \textbf{Lift vs par.} \\
      \midrule
      GPT-5.4 & \ci{64.7}{8.3} & \ci{52.6}{8.7} & \ci{58.6}{8.3} & \lift{+12.1}{7.6} & \lift{+6.1}{7.5} \\
      Gemini 3 Flash & \ci{57.8}{8.6} & \ci{47.7}{8.7} & \ci{53.8}{8.5} & \lift{+10.1}{7.2} & \lift{+4.0}{5.6} \\
      Gemini 3.1 Pro & \ci{63.9}{7.9} & \ci{55.6}{8.3} & \ci{54.9}{8.3} & \lift{+8.3}{7.9} & \lift{+9.0}{7.5} \\
      \bottomrule
    \end{tabular}
  \end{subtable}

  \vspace{0.75em}
  \begin{subtable}{\linewidth}
    \centering
    \caption{\textsc{APEX-Agents}: read-axis ablation. Mean rubric score is shown as point $\pm$ task-bootstrap 95\% half-width.}
    \label{tab:dual-artifact-ablation}
    \resulttablesetup
    \begin{tabular}{lrrrrr}
      \toprule
      \textbf{Model} & \textbf{Dual} & \textbf{Artifact-only} & \textbf{Parsed-only} & \textbf{Lift vs art.} & \textbf{Lift vs par.} \\
      \midrule
      Gemini 3 Flash High & \ci{47.8}{2.4} & \ci{43.8}{4.1} & \ci{43.1}{4.4} & \lift{+4.0}{3.4} & \lift{+4.7}{3.2} \\
      GPT-5.4 Mini xHigh & \ci{44.3}{4.4} & \ci{40.3}{4.4} & \ci{38.8}{4.0} & \lift{+4.0}{3.7} & \lift{+5.5}{3.9} \\
      GPT-5.4 Nano xHigh & \ci{42.1}{4.1} & \ci{41.1}{4.4} & \ci{32.9}{5.2} & \lift{+1.0}{3.9} & \lift{+9.2}{5.9} \\
      \bottomrule
    \end{tabular}
  \end{subtable}

  \vspace{0.75em}
  \begin{subtable}{\linewidth}
    \centering
    \caption{\textsc{APEX-Agents}: review-axis ablation on 57 file-editing tasks. Both arms use dual views and identical file tracking; only agent-visible diff tools change.}
    \label{tab:apex-diff-affordance}
    \resulttablesetup
    \begin{tabular}{@{}lrrrr@{}}
      \toprule
      \textbf{Model} & \textbf{Diffs hidden} & \textbf{Diffs visible} & \textbf{Lift} \\
      \midrule
      Gemini 3 Flash High & 47.6 & 50.2 & +2.5 \\
      GPT-5.4 Mini xHigh & 40.9 & 49.4 & +8.5 \\
      GPT-5.4 Nano xHigh & 48.1 & 51.9 & +3.8 \\
      \bottomrule
    \end{tabular}
  \end{subtable}

\end{table}

On OfficeQA (Table~\ref{tab:read-axis-summary}), the largest gap is between dual and artifact-only. Artifact-only agents can open PDFs but lack indexed evidence localization, so they spend more calls scanning broad files and are more likely to read the wrong table region. Dual improves Pass@1 by 8.3 to 12.1 points over artifact-only, with paired bootstrap significance for all three models. Parsed-only is closer to dual because the task is mostly find-and-read over a fixed PDF archive; the native view still helps on layout-heavy or fallback cases, but parsed search is the load-bearing mechanism.

On APEX (Table~\ref{tab:dual-artifact-ablation}), the pattern reverses. Parsed-only agents can find relevant instructions but cannot reliably execute on the native files. Dual improves mean rubric score by 4.7 to 9.2 points over parsed-only for all three mechanism models. The lift is largest for GPT-5.4 Nano, which is consistent with smaller models benefiting more from a synchronized workspace in this setting. Against artifact-only, paired lifts are positive for every model but smaller, which is consistent with APEX requiring both evidence localization and native execution.
The read-axis results split cleanly by task regime. OfficeQA depends most on evidence localization, while APEX also requires execution on the file state that will be graded.

The paired bootstrap tests are consistent with this interpretation. OfficeQA dual vs.\ artifact-only is significant for all three models ($p_{\mathrm{boot}}{=}0.005$ for GPT-5.4, $0.010$ for Gemini~3 Flash, and $0.041$ for Gemini~3.1 Pro); Gemini~3.1 Pro is also significant vs.\ parsed-only ($p_{\mathrm{boot}}{=}0.018$). On APEX, dual vs.\ parsed-only is significant for all three mechanism models ($p_{\mathrm{boot}}{=}2{\times}10^{-4}$ for Nano, $0.014$ for Mini, and $0.028$ for Flash). APEX dual vs.\ artifact-only has positive paired lifts for every model, but the paired-bootstrap $p$ values remain above 0.05.

Appendix Table~\ref{tab:ablation-paired-significance} shows the same pattern as paired win counts. OfficeQA has many more dual wins than artifact-only wins for GPT-5.4 and Gemini~3~Flash, while APEX has its clearest discrete advantage against parsed-only. The artifact-only comparison on APEX is closer: dual wins slightly more often for Gemini~3~Flash and GPT-5.4 Mini, but not enough to separate the arms under the secondary sign-style checks.

In the review-axis ablation, exposing workspace diffs raises scores on APEX file-editing subset by 2.5 to 8.5 points (Table~\ref{tab:apex-diff-affordance}). Both review-axis arms record the same file changes; only the visible arm exposes \texttt{workspace\_diff} and \texttt{workspace\_file\_diff} before submission. The subset is small, roughly 12\% of APEX, so it does not explain the full benchmark gains; it isolates a separate review mechanism for tasks where submitted files are modified.

The lift is largest for GPT-5.4 Mini (+8.5) and smallest for Gemini~3~Flash High (+2.5), so this slice suggests that weaker editors may benefit more from a final diff. A workspace diff does not prove semantic correctness, but it gives the model a compact chance to compare intended and actual edits before handoff.
The review-axis result points to final inspection as a supporting mechanism, not the whole system.

The narrow claim is that synchronized artifact state matters when the same agent loop must search, read, edit, and submit files. OfficeQA gives the retrieval-side case: dual access beats artifact-only on full-corpus find/read tasks. HarFeast gives the APEX-side case: parsed search identifies the survey-column guide, and native execution applies it to the same workbook that is later graded (Figure~\ref{fig:harfeast-mechanism}).

\subsection{Efficiency Diagnostics}
\label{sec:efficiency-results}

\begin{figure}[!htbp]
  \centering

  \begin{subfigure}[t]{0.49\linewidth}
    \centering
    \includegraphics[width=\linewidth]
      {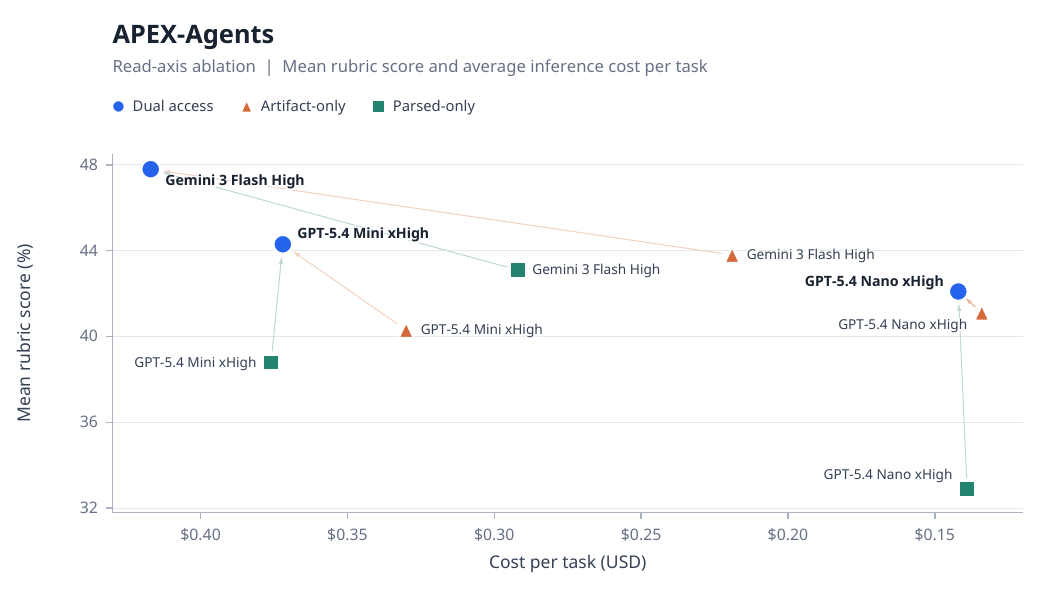}
    \caption{\textsc{APEX-Agents}: cost vs.\ Pass@1.}
    \label{fig:cost-score-officeqa}
  \end{subfigure}
  \hfill
  \begin{subfigure}[t]{0.49\linewidth}
    \centering
    \includegraphics[width=\linewidth]
      {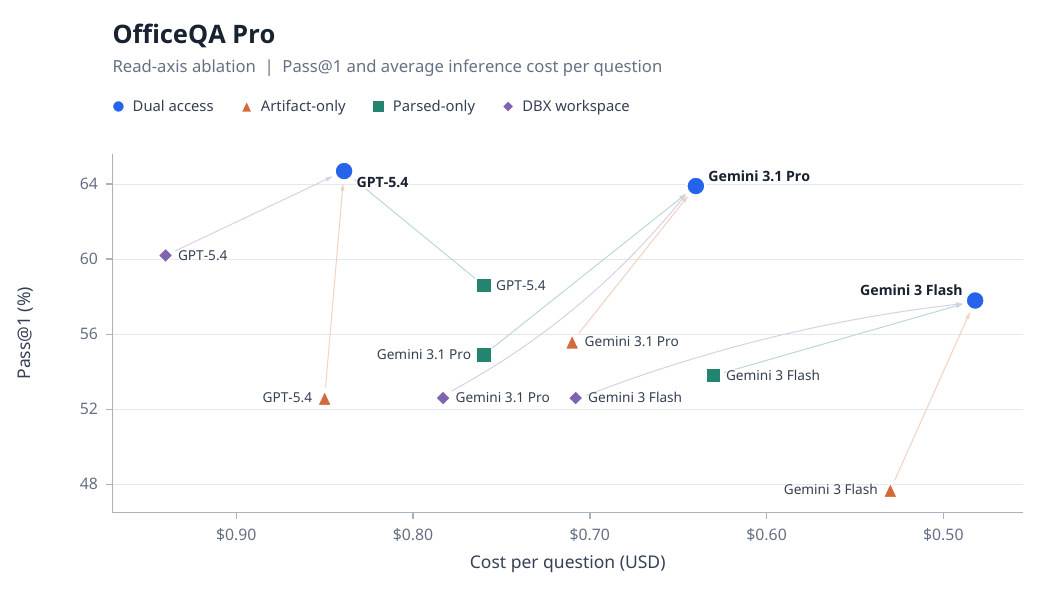}
    \caption{\textsc{OfficeQA Pro}: cost vs.\ mean score.}
    \label{fig:cost-score-apex}
  \end{subfigure}

  \caption{Diagnostic cost--score ablations. Lower and further right is better: a run scores more while spending less. Each arrow points from an ablated access configuration to the dual-access result for the same model. Dual access generally improves diagnostic score, although the associated cost change varies across models and benchmarks.}
  \label{fig:cost-score-ablation}
\end{figure}

Accuracy does not rise simply because the dual runs are longer or more expensive. In OfficeQA, Figure~\ref{fig:cost-score-ablation} places the three dual rows on the upper-right frontier: each dual row has the highest score for its model, and for Gemini~3 Flash and Gemini~3.1 Pro it is also the lowest-cost row. GPT-5.4 dual is nearly cost-matched with artifact-only (\$0.84 vs.\ \$0.85 per question) and costs less than the Databricks parsed-only baseline (\$0.94), while improving Pass@1 over both. Dual is also faster than Reducto parsed-only for all three OfficeQA models: 6.1 vs.\ 17.3 minutes for GPT-5.4, 15.3 vs.\ 18.8 minutes for Gemini~3 Flash, and 10.3 vs.\ 18.7 minutes for Gemini~3.1 Pro.

The APEX panel shows a different tradeoff. Dual is the highest-scoring diagnostic row for Gemini~3 Flash and GPT-5.4 Mini, and for Mini it is also faster than both restricted views (7.7 minutes vs.\ 8.8 for artifact-only and 10.3 for parsed-only). For GPT-5.4 Nano, artifact-only is a small diagnostic cost-score exception, but dual still avoids the parsed-only problem at essentially the same agent-run cost. The Pareto plots therefore support the mechanism results rather than replacing them: the APEX gains sometimes trade some cost for better execution on native files, whereas the OfficeQA gains do not.

% \begin{figure}[!htbp]
%   \centering
%   \includegraphics[width=0.76\linewidth]{Figs/cost_score_pareto_officeqa.pdf}\\[-0.35em]
%   \footnotesize (a) \textsc{OfficeQA Pro}: cost vs.\ Pass@1.\\[0.5em]
%   \includegraphics[width=0.76\linewidth]{Figs/cost_score_pareto_apex.pdf}\\[-0.35em]
%   \footnotesize (b) \textsc{APEX-Agents}: cost vs.\ mean score.
%   \caption{Diagnostic cost-score Pareto panels. Higher and further left is better: a run scores more while spending less. In OfficeQA, dual forms the upper-left skyline over the single-view arms. In APEX, dual gives the highest diagnostic score for GPT-5.4 Mini and Gemini~3 Flash, while GPT-5.4 Nano shows a small artifact-only cost-score exception; parsed-only remains below its paired dual row.}
%   \label{fig:cost-score-pareto}
% \end{figure}

% Preamble

\subsection{Mechanism and Error Analysis}
\label{sec:mechanisms}
\label{sec:residual-failures}

Trajectory review supports the benchmark-level pattern. In OfficeQA, dual wins over artifact-only; most gains come from the parsed view; visual fallback matters less often but helps on layout-heavy scans. In APEX, dual often uses the parsed view to find instructions and the native view to execute on deliverables.

\begin{figure}[!htbp]
  \centering
  \includegraphics[width=\linewidth]{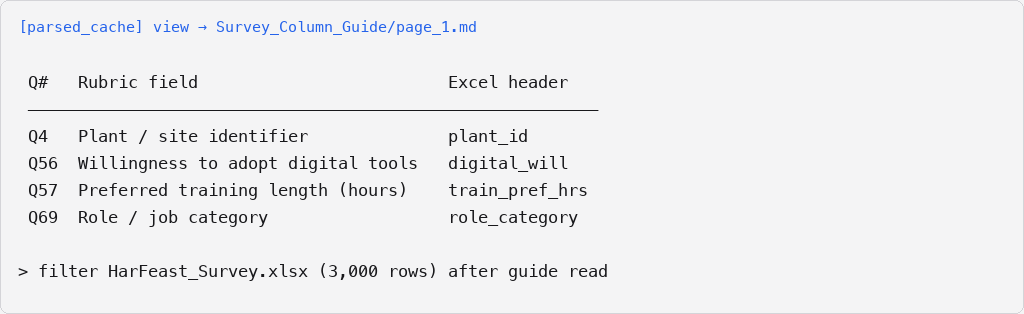}
  \caption{\textsc{APEX-Agents} HarFeast trace. The parsed cache identifies the relevant survey-column guide, and native workbook execution applies that interpretation to the same 3{,}000-row spreadsheet that is later graded. The example illustrates the APEX mechanism: evidence localization and native execution must refer to the same workspace state.}
  \label{fig:harfeast-mechanism}
\end{figure}

A concrete HarFeast workforce task shows the APEX mechanism at the trace level. The agent had to interpret a survey column guide and then filter a 3{,}000-row workbook. In the dual arm, parsed search exposed the relevant guide text, and the subsequent \texttt{pandas} operation ran on the same hydrated workbook that would be graded. Artifact-only could open the workbook, but miscomposed the filtering logic. Parsed-only could find instructions but could not execute on the native workbook.
Figure~\ref{fig:harfeast-mechanism} shows why the two views need to be synchronized: the parsed evidence identifies the column semantics, while native execution applies that interpretation to the workbook used for grading.

Across dual-win model-task instances, parsed-cache use and native-file execution account for most coded mechanisms (Table~\ref{tab:apex-dual-mechanisms}).

The remaining failures show the limits of workspace synchronization. OfficeQA losses often involve subtle numerical reasoning, ambiguous table scope, or a correct source followed by an incorrect computation. On APEX, 188 of 452 evaluated tasks are all-arm zeros for at least one completed model triplet, with Investment Banking accounting for the largest share (Table~\ref{tab:residual-error-analysis}). These failures are not primarily state-sync failures; they require stronger planning, domain reasoning, validation, or task-specific tooling.

\FloatBarrier

\section{Discussion}
\label{sec:discussion}

The paired ablations isolate the effects of artifact access and review visibility within this workspace design. The failure is mundane but consequential: an agent can reason over one file state and submit another. The claim is not that agents simply need more tools; removing a view changes performance in the direction predicted by each task mechanism. Large all-arm-zero residuals show that artifact-state synchronization does not remove planning, domain reasoning, or rubric-following failures.

The size of the read-axis effect depends on the base model: non-frontier models gain most (Gemini 3.1 Pro rises from 29.3\% to 63.9\% on OfficeQA), while frontier models change little.
We read this as evidence of a possible interaction between artifact-state access and model capability: in these runs, a synchronized workspace recovers some of the gap that a larger model may otherwise compensate for through repeated search and re-reading.
A ceiling effect for the strongest models remains possible, so we report this as an interaction rather than a clean substitution.
Either way, a benchmark that leaves the workspace uncontrolled can mix model capability with workspace-state access.
Because the two interact in our ablations, cross-harness leaderboard gaps may overstate the capability difference between small and large models on knowledge-work tasks.

\paragraph{Limitations.}
We designed the study to control the main confounds, and three limits remain.
First, the benchmark graders score only the final answer or materialized deliverable, not whether retrieval, workspace sync, or diff visibility succeeded at intermediate checkpoints. Neither rubric scores trajectories or checkpointed mini-successes (e.g., opening the correct source, reading the correct table region, or making a valid tracked edit). To recover part of that signal, we hand-code trajectories (Sections~\ref{sec:mechanisms} and~\ref{sec:residual-failures}); this labeling is illustrative rather than exhaustive and can vary across coders, so we read it as supporting evidence rather than a verifier on the level of executable patch tests (Section~\ref{sec:relwork}).
Second, the published comparison rows are not one-to-one, since they differ in attempt budget ($^{\dag}$), agent harness, retrieval backend, and parser choice. We therefore rest every causal read-axis claim on paired three-arm ablations that hold the model, prompt, parser, retriever, grader, file tracker, and tool budget fixed. A parser-sensitivity check adds support: \emph{parsed-only} Pass@1 shifts by ${<}2$\,pp across two exports of the same Treasury PDFs (Table~\ref{tab:read-axis-summary}). These ablations still hold the parser and retriever fixed, so they do not yet separate the workspace-state contract from those two components. Because existing benchmarks score final answers or deliverables rather than checkpointed state transitions, our ablations test the separable artifact-view and review-view effects available under these interfaces; they do not establish that the proposed state model is the unique minimal abstraction.

Third, synchronization does not remove every failure: $188/452$ evaluated \textsc{APEX} tasks are all-arm zeros for at least one model (Table~\ref{tab:residual-error-analysis}) and do not move under read-axis toggles, which marks where state-management gains end and planning, domain-reasoning, and rubric-following errors begin (Section~\ref{sec:residual-failures}).

\paragraph{Future Directions.}
Workspace state is a systems question, not only a model-scaling or prompt-tuning question. Several tests remain open.
Our ablations fix one agent stack, parser, retriever, and prompt, so a direct next step is to vary these components. That means testing transfer across parsers, retrievers, prompts, and human-in-the-loop review policies, and running component-level ablations that separate the workspace-state contract from the parser and retriever themselves.
The review-axis result rests on the ${\approx}12\%$ file-editing slice of \textsc{APEX} (Section~\ref{sec:diff-ablation}), so a larger edit-heavy evaluation would test how dual views interact with diff visibility and cell-level edit tools.
On the benchmark side, future suites should expose which state is evaluated (accepted files, parsed records, or staged edits) rather than only terminal success. They should also move toward software-engineering-style acceptance criteria, including patch-and-test contracts, trajectory analysis, and verifiers adapted to retrieval, read-region, and edit checks rather than inferred from traces alone.

\section{Conclusion}
\label{sec:conclusion}

Long-horizon knowledge-work agents must find sources, read layout-heavy evidence, edit deliverables, and produce reviewable outputs over many turns.
Reliability can break when parsed cache, semantic search, and final artifacts refer to different versions of the same files. We formalize synced retrieval, editing, and review through dual artifact views, hash-keyed parsed-index refresh, and diff exposure on edits. Fixed-harness ablations on \textsc{OfficeQA Pro} and \textsc{APEX-Agents} provide evidence that this versioned-workspace invariant can improve read-axis performance in these benchmarks and help on a small edited-deliverable slice; the efficiency diagnostics indicate that these gains are not simply higher-cost runs. Dual views do not eliminate reasoning, convention, or tool-use failures; a large \textsc{APEX} residual is unchanged by read-axis toggles.
Outcome-only rubrics reveal whether an answer passed, not whether retrieval, reading, or edit validity failed.
Next-generation knowledge-work agent benchmarks should pair work requests with inspectable acceptance criteria over evidence, staged edits, and final work products.

\section*{Acknowledgments and Disclosure of Funding}
Funding in direct support of this work: Reducto provided in-kind API credits used to instantiate the document parser in our \systemshort{} deployment.

\bibliographystyle{abbrvnat}
\bibliography{references}

\appendix
\setcounter{table}{0}
\renewcommand{\thetable}{A.\arabic{table}}
\setcounter{figure}{0}
\renewcommand{\thefigure}{A.\arabic{figure}}

\section{Implementation Details}
\label{app:implementation-details}

The workspace-state model (Figure~\ref{fig:artifactstate-architecture}) and agent turn loop (Section~\ref{sec:eval-agent-turn}) are defined in Section~\ref{sec:substrate}; this appendix records operational specifics and update rules omitted there.

\paragraph{Commit and Promotion Semantics.}
\label{app:substrate-formal}
Let $t$ index the current agent turn. 

The implementation keeps an accepted file tree and an ordered journal of proposed file operations, but the agent-facing state remains the three-view workspace from Section~\ref{sec:artifactstate-contract}: native files $W_t$, parsed cache $C_t$, and review diff $\Delta_t$.
Each file operation records the actor, target path, base hash, operation type, and payload.
If the recorded base hash no longer matches the current accepted hash of the target path, the operation is treated as stale and left for agent or reviewer reconciliation.
Non-conflicting operations remain in journal order and define the current native workspace $W_t$.
If verification passes and the edit is accepted, those operations are promoted to the accepted file tree; otherwise they remain staged.
After promotion or staging changes, the parsed cache is recomputed or marked stale for changed files before the next search call.

The diff surface is format-aware: line diffs for text, document previews for word-processing files, row- and cell-level spreadsheet views, slide-level presentation views, and before/after previews for binary artifacts.

\paragraph{Versioned Workspace and Branches.}
The workspace is a tree of directories and artifacts with per-user, kind-scoped branches (chat, review, evaluation).
Each agent turn creates a turn-scoped patch; human edits use lazily created human patches on the same branch chain.
The reducer maps the accepted tree and staged operations to $W_t$, diffs, a commit plan, and conflicts; commit is transactional and rollback uses inverse patches.
Citations resolve to artifact addresses and page numbers shared by diffs, action items, and chat.

\paragraph{Parsing, Retrieval, and Sandbox Sync.}
Section~\ref{sec:setup-harness} states the main implementation contract: hydrated sandbox files are native access, and Reducto-derived records and search are parsed access.
Parser-generated companions and in-sandbox parse outputs are excluded from the journal and regenerated on commit.
Format-specific skill packs for spreadsheets, documents, and slides run inside the sandbox rather than on the first-class tool surface.
The benchmark corpora exercise hash-keyed reuse at two file-count regimes: one shared OfficeQA workspace with ${\approx}697$ PDFs and \textsc{APEX} task folders averaging ${\approx}166$ mixed-format files.
We do not claim a separate asymptotic scalability benchmark beyond those workloads; large repositories still pay the cost of initial parse and index construction.

\section{Experimental Protocols}
\label{app:setup}

\subsection{Agent Execution Environment}
\label{app:agent-runtime}

\paragraph{Sandbox.}
Each agent turn runs in an isolated Linux sandbox (Daytona) using the \emph{small} resource profile: 1 vCPU, 1\,GiB RAM, and 3\,GiB disk.
The sandbox is created on first tool use in a turn.
Shell and Python tools use a 30\,s default timeout and a 600\,s maximum.

\paragraph{Tools.}
Benchmark agents share the standard tool surface: sandbox execution (\texttt{bash}, \texttt{python}), file inspection (\texttt{view}, \texttt{list\_dir}), workspace search (\texttt{grep}, \texttt{file\_search}, \texttt{vector\_search}), tracked-change inspection (\texttt{workspace\_diff}, \texttt{workspace\_file\_diff}), light file edits, session notes (\texttt{read\_whiteboard}, \texttt{write\_whiteboard}), and, on the OfficeQA internet subset only, \texttt{web\_search} and \texttt{web\_fetch}.
Artifact-view arms disable parsed search (\emph{artifact-only}) or remove originals from the sandbox (\emph{parsed-only}) as in Section~\ref{sec:setup-ablation}.
All runs share the 250-tool-call budget from Section~\ref{sec:setup-benchmarks}.

\paragraph{Agent Prompting.}
Every read-axis arm uses the same base agent system prompt, so any score movement is attributable to artifact views rather than to instruction wording.
The prompt states the agent's role, the workspace conventions (paths under \texttt{/root/workspace/}, parsed companions, and freshness labels), the tool surface in Table~\ref{tab:agent-tool-routing}, and each benchmark's submission rule; the same text is used in all three arms even though the available tools differ.
The verbatim agent prompts ship with the sanitized implementation (Appendix~\ref{app:responsible-release}).

\paragraph{Agent-Visible Reads, Search, and Parse Freshness.}
\label{app:agent-artifact-layer}
On first tool use the sandbox hydrates $W_t$ under \texttt{/root/workspace/}: each artifact (e.g.\ \texttt{report.pdf}) plus a parsed companion tree (\texttt{report.pdf\_parsed/page\_N.md}) when $C_t$ is enabled.
Reducto output is per-page markdown with body text, HTML tables for complex layouts, \texttt{media/} figure paths, and optional agentic captions, chunked for \texttt{vector\_search}.

Text views on a parseable file return parsed pages with metadata for source, parse time, source update time, and freshness. A \texttt{stale} freshness value means the source changed before re-parse could finish, so the agent should prefer source files or wait for re-indexing.
When $C_t$ is disabled or missing, \texttt{view} labels output as \texttt{pdftotext} or Office fallback.
Visual file views render selected PDF or Office pages to images for layout, charts, and scans; they are available whenever originals are in the sandbox (dual and \emph{artifact-only}).
\texttt{grep} on a PDF path resolves to parsed page files and returns \texttt{path:page} locators; without $C_t$, content search over legacy scans is often empty and agents fall back to shell extraction (Section~\ref{sec:mechanisms}).
\texttt{python}/\texttt{bash} operate on native files (\texttt{openpyxl}, etc.) and drive \textsc{APEX} deliverable edits.

\begin{table}[H]
  \caption{Read-path tool routing by artifact-view arm.}
  \label{tab:agent-tool-routing}
  \centering
  \appendixtablesetup
  \begin{tabularx}{\linewidth}{@{}lXXX@{}}
    \toprule
    \textbf{Tool} & \textbf{Dual} & \textbf{Artifact-only} & \textbf{Parsed-only} \\
    \midrule
    \texttt{view(text)} & Parsed pages with freshness metadata & \texttt{pdftotext}/Office fallback & Parsed pages only \\
    \texttt{view(visual)} & Native render & Native render & Unavailable \\
    \texttt{grep}/\texttt{file\_search} & Parsed page locators & Raw paths only; weak for PDFs & Parsed page locators \\
    \texttt{vector\_search} & Page-chunk embeddings & Disabled & Page-chunk embeddings \\
    \texttt{python}/\texttt{bash} & Native bytes and deliverables & Native bytes and deliverables & No native deliverables \\
    \bottomrule
  \end{tabularx}
\end{table}

\paragraph{\textsc{OfficeQA Pro} Databricks \texttt{.txt} Baseline vs.\ Live Parser.}
\citet{officeqa2026} evaluate agents on a pre-exported \texttt{.txt} corpus: parsed elements from Databricks \texttt{ai\_parse\_document} concatenated in reading order, with figures omitted.
Our primary \emph{parsed-only} and dual arms instead use live Reducto parsing over hydrated PDFs, hash-keyed re-indexing, and the tool routes in Table~\ref{tab:agent-tool-routing}; the Databricks \texttt{.txt} baseline is a leaderboard-aligned comparison.

\paragraph{Replication Pointers.}
Graders and task definitions follow the benchmark protocols in Section~\ref{sec:setup-benchmarks}; baseline rows in headline tables are quoted without recomputation.
The grader system prompts and protocols are provided in Appendix~\ref{app:grader-protocols}.

\subsection{Grading Protocols and Grader Prompts}
\label{app:grader-protocols}

Our \systemshort{} eval harness runs a separate grader LLM session after each agent attempt.
The grader user message always includes \texttt{<ORIGINAL\_TASK>} (case prompt plus optional run extra prompt), \texttt{<TEXT\_RESPONSE>} (final assistant message), optional \texttt{<ARTIFACTS>} (workspace-relative paths of files the agent created or modified versus the run baseline; omitted when empty), and \texttt{<RUBRIC>} (benchmark rubric items).
Grader model, decoding settings, and prompt version are fixed per run and identical across artifact-view ablations (Table~\ref{tab:ablation-design}); only the agent artifact-view arm varies.

\paragraph{\textsc{OfficeQA Pro}.}
\label{app:grader-officeqa}
Each question has one answer-match rubric item with the gold string embedded in the rubric text.
The grader may call only \texttt{submit\_eval\_grade} (no file tools).
Scoring follows deterministic exact-match semantics aligned with the public benchmark's \texttt{reward.py} convention: 0.0\% allowable absolute relative error, with no partial credit~\citep{officeqa2026}.
Pass@1 is the fraction of questions with grader \texttt{score}${=}1$.
The system prompt is given verbatim below.

\begin{Verbatim}
You are evaluating answer correctness for a Document QA benchmark.

# Your Computer
You have access to only one tool: `submit_eval_grade`. You are not allowed to use any other tool.

# Numeric Matching Rules
When checking numeric answers:
- Compare based on value, not presentation.
- Accept equivalent renderings such as:
  - 1,250 = 1250
  - $1,250 = 1250
  - 0.25 = 25%
  - 3.500 = 3.5
- Reject:
  - rounded answers when the exact value is expected,
  - answers with the wrong unit or scale,
  - ranges when a single value is expected,
  - values that require unstated assumptions to be considered correct.
- If a unit is required by the criterion, the unit must also be correct.

# Instructions
Compare the agent's final answer to the expected answer.

- Mark correct only if they match exactly in meaning.
- If the expected answer contains multiple values (e.g. [a, b]), all required values must be present and correct.
- Order matters unless the Evaluation Criteria explicitly states otherwise.
- If the response contains conflicting or inconsistent values for the same answer, mark incorrect.
- For numeric answers, require exact equivalence under the Numeric Matching Rules (no approximation or "close enough").
- Ignore formatting differences (commas, symbols, whitespace) but not value, units, or scale.
- Partial correctness counts as incorrect.

After deciding, call `submit_eval_grade` with `score = 1.0` only for fully correct answers; otherwise `score = 0.0`.
\end{Verbatim}

\paragraph{\textsc{APEX-Agents}.}
\label{app:grader-apex}
Each task has multiple pass/fail rubric criteria (Section~\ref{sec:setup-benchmarks}).
The grader runs \texttt{google/gemini-3-flash-preview} with thinking level \emph{low}, matching~\citet{apexagents2026}.
The grader may call \texttt{view} on paths listed in \texttt{<ARTIFACTS>} (or paths named in the task) and must return one \texttt{submit\_eval\_grade} with a score per rubric id; a task passes only if every criterion receives \texttt{score}${=}1$ on that attempt.
The system prompt is given verbatim below.

\begin{Verbatim}
You are an expert evaluator grading an AI agent's work. Determine if each rubric criterion was met from the agent's final response and/or changed files. Be precise, evidence-based, and objective.

<GRADING_PRINCIPLES>
- Grade only what that criterion asks--nothing more, nothing less.
- Do not penalize for issues the criterion does not mention.
- Use evidence allowed for that criterion (see <TASK_SCOPE>); infer chat vs file grading from <ORIGINAL_TASK> plus that criterion's wording.
- Be consistent; do not fabricate file contents or facts that do not appear in allowed evidence.
</GRADING_PRINCIPLES>

<YOUR-COMPUTER>
Tools: `view` (read `/root/workspace/`; no writes) and `submit_eval_grade`. For heavy spreadsheets/documents, `view` with `mode="visual"` or read `/root/skills/<docx|pdf|pptx|spreadsheet>/SKILL.md` via `view`.
</YOUR-COMPUTER>

<TASK_SCOPE>
1. The user message contains <ORIGINAL_TASK>, <TEXT_RESPONSE>, optional <ARTIFACTS>, and <RUBRIC>. You do not receive the rest of the conversation.

2. Deliverable first: If <ORIGINAL_TASK> requires creating or editing a file or artifact, rubric items about that deliverable's existence or contents must be graded from the actual file via `view` (paths in <ARTIFACTS> and/or named in the task). Chat alone is insufficient for those. If the deliverable cannot be identified from the task + <ARTIFACTS>, do not guess--fail that file-related criterion (or "uncertain" only if the rubric allows).

3. Answer-only: If the task does not require producing or changing a deliverable--only a reply in chat--then rubric items about what the agent said, concluded, mentioned, or explained are graded only from <TEXT_RESPONSE>. Do not fail those items because you did not `view` source materials.

4. Hybrid: If the task requires both a deliverable and chat, decide per rubric item: file/content checks -> `view` the artifact; chat wording or reasoning called for in that item -> <TEXT_RESPONSE>.

5. "Do not infer": Do not invent text, numbers, or file contents that are not in allowed evidence. It does not mean "reject <TEXT_RESPONSE> unless files were opened" for items graded from chat. For file items, missing paths or ambiguous artifact identity -> fail if the rubric does not allow uncertainty.
</TASK_SCOPE>

<VIEW_USAGE>
- Call `view` when the criterion requires file/artifact content, or when <ARTIFACTS> lists paths and the criterion is about those changes.
- Use only paths inside <ARTIFACTS> (or paths the task explicitly names as the deliverable). Do not derive paths from `file://` or other citations in the reply; do not guess filenames.
- If file content is needed for a criterion but no allowed path exists, you cannot verify from disk--grade that item from <TEXT_RESPONSE> only when the criterion is purely about chat coverage; otherwise fail (or uncertain if allowed).
</VIEW_USAGE>

<ARTIFACT_RULES>
- For file-change criteria, evidence is inspected content at <ARTIFACTS> paths only; chat claims like "I updated the file" are not proof.
- If there is no <ARTIFACTS> section, treat as no staged file changes: criteria that require edited/new file proof fail; criteria that only judge chat may still pass.
- Do not hallucinate file contents--open with `view` when the criterion depends on artifact content.
- If the agent claims edits but <ARTIFACTS> does not support a required deliverable, file-dependent criteria fail.
</ARTIFACT_RULES>

<EVALUATION_STANDARD>
Two kinds (check which applies to this criterion):

(i) Chat coverage -- Criterion asks whether the response states, concludes, finds, mentions, explains, or identifies something:
Pass if <TEXT_RESPONSE> clearly conveys the required position. Reasonable paraphrase and careful/qualified phrasing count unless the criterion demands exact terms. Nuance in another paragraph does not fail the item unless it contradicts the required claim on that point. If the criterion also requires the reasoning to align with the conclusion, fail when the explanation contradicts the conclusion for that claim.

(ii) Verified facts / files / numbers -- Everything else (including artifact correctness):
Match closely--partial or wrong details fail unless <TOLERANCE_RULES> apply.

Shared rules:
- When file or artifact proof is required, use `view` only on allowed paths from <ARTIFACT_RULES> / <VIEW_USAGE>.
- Conjunctive "X and Y": verify each part.
- Specificity: broad category -> address all required members; specific term -> a vaguer substitute may fail unless (i) clearly applies.
- File-specific: If the criterion names a specific file or type for content, only that target counts; other files are irrelevant; agent prose cannot substitute for `view` when the item is about file content; if the target is missing, fail.

</EVALUATION_STANDARD>

<TOLERANCE_RULES>
- Numeric formatting that preserves the value is OK (e.g. $153.5 vs $153.50; 10 vs 10.0).
- Values that round to the precision stated in the criterion are OK (e.g. $2.07B for "$2.1bn"; $26.83B for "$26.8bn"); use the criterion's rounding rule if it gives one.
- Treat .xls/.xlsx, .doc/.docx, .ppt/.pptx as the same format when matching.
</TOLERANCE_RULES>

<RATIONALE_FORMAT>
Each `note`: Evidence (1-3 short quotes or file/location refs) and Assessment (quote the criterion; pass/fail with why). Under ~400 words. For chat-graded items, quote <TEXT_RESPONSE>; for file items, cite path + location (sheet/row/section).
</RATIONALE_FORMAT>

<OUTPUT_FORMAT>
Call `submit_eval_grade` exactly once: `{ results: [{ id, score: 0|1, note? }] }` for every rubric id, then stop.
Each `note` must follow <RATIONALE_FORMAT>.
</OUTPUT_FORMAT>
\end{Verbatim}

\ifdefined\artifactstatepreprint
  \FloatBarrier
\fi

\subsection{Efficiency Protocols}
\label{app:setup-cost}
\label{sec:setup-cost}

\paragraph{Pass@$k$ and Multi-Attempt Rates (GPT Family, Three Attempts).}
Table~\ref{tab:apex-pass-at-k} summarizes the same GPT-family dual runs as Section~\ref{sec:setup-metrics}.
Our headline APEX Pass@1 follows the benchmark's per-attempt reporting convention over three reruns per task, not the fraction of tasks passed on at least one attempt.
Pass$^{\times 3}$ means all three attempts pass; Pass@any means at least one of the three attempts passes.
The public \textsc{APEX-Agents} leaderboard reports Pass@1 with eight independent attempts per task (Section~\ref{sec:setup-metrics}); we quote those cells as published (\apexldb).

\begin{table}[!htbp]
  \centering
  \resulttablesetup
  \caption{\textsc{APEX-Agents}: Historical multi-attempt diagnostic for GPT-family dual runs. \textbf{Pass@1} is the mean per-attempt pass rate in that diagnostic. \textbf{Pass$^{\times 3}$}: all three attempts pass. \textbf{Pass@any}: at least one passing attempt. This diagnostic is not used for the main paired ablations.}
  \label{tab:apex-pass-at-k}
  \begin{tabular}{lrrr}
    \toprule
    Model & Pass@1 (\%) & Pass$^{\times 3}$ (\%) & Pass@any (\%) \\
    \midrule
        GPT-5.4 & 37.7 & 29.0 & 47.3 \\
        GPT-5.4 Mini & 27.7 & 16.5 & 39.6 \\
        GPT-5.4 Nano & 25.0 & 14.6 & 35.9 \\
    \bottomrule
  \end{tabular}
\end{table}

\paragraph{Cost and Latency.}
\textbf{Cost} per graded completion is mean logged \emph{agent-run} spend: we sum \texttt{cost\_micros} on assistant and tool-call turns in the eval session at public list rates for the model and run date.
Background workspace parsing (Reducto credits, Mistral OCR pages) and Daytona sandbox host CPU are recorded as separate internal usage and are \textbf{not} included in the cost columns in Tables~\ref{tab:officeqa-ablation-efficiency} and~\ref{tab:apex-sw-efficiency}.
Dual and \emph{parsed-only} arms therefore omit one-time ingest cost for the search cache; reported costs compare in-run model spend across artifact-view arms on the same workspace.
For the efficiency tables and Pareto figures, \textbf{latency} is the mean agent time per graded completion: wall-clock time from agent start to agent finish minus sandbox startup time recorded on the chat turn.
Startup time is excluded from the reported latency column, but it still constrains when tools can return; configurations with longer per-case setup times therefore see higher end-to-end wall times than the table latency alone suggests.
Tool-call counts and cost use the full graded completion.
Means are taken over graded completions in each configuration on the benchmark pools defined in Section~\ref{sec:setup-benchmarks}.
Published OfficeQA agent rows are descriptive leaderboard metrics reported in Table~\ref{tab:complete-system-results}.

\paragraph{Ablation Table Alignment.}
Tables~\ref{tab:read-axis-summary} and~\ref{tab:dual-artifact-ablation} are the source of the primary paired read-axis estimates: OfficeQA Pass@1 lifts and APEX mean rubric-score lifts with bootstrap intervals as defined in Section~\ref{sec:setup-metrics}.
The cost-score panels and efficiency tables are diagnostic surfaces: they use the logged latency, cost, and tool-count export and should not be read as a replacement for the paired intervals in Tables~\ref{tab:read-axis-summary} and~\ref{tab:dual-artifact-ablation}.
The OfficeQA efficiency table lists all OfficeQA artifact-view arms with latency, cost, and tool counts.
Table~\ref{tab:officeqa-reported-comparison} quotes published \textsc{OfficeQA Pro} Full and Oracle agent rows alongside our dual runs on matched models.
Table~\ref{tab:apex-reported-comparison} quotes published \textsc{APEX-Agents} leaderboard rows alongside our dual runs on matched models.
OfficeQA has 133 questions; APEX has 480 tasks, of which our evaluated pool contains 452.
For the APEX baseline, we report published cost or latency values when available and mark unavailable cells N/A.

\paragraph{Paired Discordance Checks on Ablation Triplets.}
Tables~\ref{tab:read-axis-summary} and~\ref{tab:dual-artifact-ablation} report paired $\Delta$ values from the bootstrap contrasts. Section~\ref{sec:ablation} reports the two-sided $p_{\mathrm{boot}}$ values, and Table~\ref{tab:ablation-paired-significance} adds McNemar/sign tests on per-task win counts as secondary checks.
\textbf{Dual wins} and \textbf{Restr.\ wins} count tasks where dual beat the restricted arm named in \textbf{Contrast}, or the reverse (ties excluded).
On \textsc{APEX}, all three models have $p_{\mathrm{boot}}{<}0.05$ vs.\ \emph{parsed-only}; none do vs.\ \emph{artifact-only} (e.g., Gemini~3 Flash: 68 dual wins vs.\ 65 restricted wins, $p_{\mathrm{boot}}{=}0.23$).
Every evaluated model still has at least one row with $p_{\mathrm{boot}}{<}0.05$ (from \textsc{OfficeQA} and/or \textsc{APEX} vs.\ \emph{parsed-only}).

\begin{table}[!htbp]
  \centering
  \appendixtablesetup
  \setlength{\tabcolsep}{2.2pt}
  \caption{Read-axis paired-discordance checks. $p_{\mathrm{disc}}$ is McNemar for OfficeQA and a sign test for APEX; these count paired wins and losses as a discrete check on the bootstrap contrasts.}
  \label{tab:ablation-paired-significance}
  \begin{tabular}{@{}lllrrr@{}}
    \toprule
    \textbf{Bench.} & \textbf{Model} & \textbf{Contrast} & \textbf{Dual wins} & \textbf{Restr.\ wins} & \textbf{$p_{\mathrm{disc}}$} \\
    \midrule
    \textsc{OfficeQA} & GPT-5.4 & vs artifact-only & 20 & 6 & \textbf{0.009} \\
    \textsc{OfficeQA} & GPT-5.4 & vs parsed-only & 16 & 10 & 0.327 \\
    \midrule
    \textsc{OfficeQA} & Gemini 3 Flash & vs artifact-only & 17 & 5 & \textbf{0.017} \\
    \textsc{OfficeQA} & Gemini 3 Flash & vs parsed-only & 9 & 5 & 0.424 \\
    \midrule
    \textsc{OfficeQA} & Gemini 3.1 Pro & vs artifact-only & 20 & 9 & 0.061 \\
    \textsc{OfficeQA} & Gemini 3.1 Pro & vs parsed-only & 19 & 7 & \textbf{0.029} \\
    \midrule
    \textsc{APEX} & Gemini 3 Flash High & vs artifact-only & 68 & 65 & 0.862 \\
    \textsc{APEX} & Gemini 3 Flash High & vs parsed-only & 85 & 60 & \textbf{0.046} \\
    \midrule
    \textsc{APEX} & GPT-5.4 Mini xHigh & vs artifact-only & 100 & 86 & 0.341 \\
    \textsc{APEX} & GPT-5.4 Mini xHigh & vs parsed-only & 107 & 87 & 0.172 \\
    \midrule
    \textsc{APEX} & GPT-5.4 Nano xHigh & vs artifact-only & 85 & 99 & 0.338 \\
    \textsc{APEX} & GPT-5.4 Nano xHigh & vs parsed-only & 110 & 79 & \textbf{0.029} \\
    \bottomrule
  \end{tabular}
\end{table}

\ifdefined\artifactstatepreprint
  \FloatBarrier
\fi

\section{Benchmark Selection}
\label{app:benchmark-selection}

Section~\ref{sec:setup-benchmarks} describes the two evaluated benchmarks, \textsc{OfficeQA Pro} and \textsc{APEX-Agents}.
Before settling on this pair, we surveyed knowledge-work agent benchmarks cited in Section~\ref{sec:relwork} against the evaluation needs of our versioned workspace: isolating whether parsed search, native reads, and edit visibility refer to the same hash-versioned working tree.
A later appendix records desiderata for future benchmark design; here we summarize inclusion criteria and why several widely used alternatives were not used as primary evaluators.

\paragraph{Inclusion Criteria.}
We required benchmarks that:
\begin{itemize}[leftmargin=1.5em,itemsep=2pt,topsep=2pt]
  \item target knowledge-work products, such as professional PDFs, spreadsheets, decks, and memos, rather than code repositories or single-page trivia QA;
  \item expose real file complexity, including large shared corpora or per-task project folders with mixed formats and layout-heavy tables;
  \item demand a long-horizon agent loop spanning retrieve, read, plan/execute, and submit, rather than one-shot parsing, layout detection, or GUI clicking alone;
  \item ship \textbf{public task pools and graders} so we do not retrofit unrelated QA sets into artificial shared-context groups; and
  \item run on our fixed \systemshort{} harness with identical prompts, tool budgets, and three-way artifact-view ablations.
\end{itemize}

\paragraph{GUI- and Application-Control Suites.}
OSWorld~\citep{xie2024osworld}, The Agent Company~\citep{xu2024theagentcompany}, and OdysseyBench~\citep{odysseybench2025} stress multi-app GUI control and long office \emph{workflows}.
They measure action reliability in real desktops, but they under-specify the read-axis question we ablate: whether the parsed cache and sandbox originals stay synchronized when the agent searches, opens, and edits files.
Failures on those leaderboards often mix UI grounding, app state, and file-state errors; our study instead holds the UI surface fixed while toggling artifact views for knowledge-work tasks.

\paragraph{Parsing- and Layout-Fidelity Benchmarks.}
DocLayNet~\citep{pfitzmann2022doclaynet}, PubTables~\citep{smock2021pubtables}, OmniDocBench~\citep{ouyang2024omnidocbench}, and ParseBench~\citep{zhang2026parsebench} score component- or page-level conversion quality.
They do not instantiate a multi-turn agent over a versioned workspace, so they cannot test whether hash-keyed re-parsing keeps search aligned with the source files even after edits.
We use a document parser inside \systemshort{}, but parser F1 is not a substitute for end-to-end artifact-view ablations.

\paragraph{Software-Engineering Benchmarks.}
SWE-bench introduced the repository-and-test interface for real-world software issues \citep{jimenez2023swebench}. SWE-bench Verified refines the task set \citep{openai2024swebenchverified}, and SWE-Bench Pro extends the setting to longer-horizon software tasks \citep{deng2025swebenchpro}. Together, these benchmarks are the closest systems analog because they use a tracked tree, derived retrieval, and executable verifiers. Their artifact surface is source code with line-addressable \texttt{grep} and patch tests, not layout-heavy office deliverables.
\textsc{APEX-SWE}~\citep{kottamasu2026apexswe} targets the same professional-work theme as \textsc{APEX-Agents} yet remains code-centric; it does not cover the read-heavy Treasury corpus regime that OfficeQA supplies.

\paragraph{Lighter Office-Automation Suites.}
OfficeBench~\citep{wang2024officebench} benchmarks multi-app office automation with shorter task chains than \textsc{APEX-Agents} and without OfficeQA's full-corpus retrieval stress (${\approx}697$ shared PDFs).
It is a useful complement, but pairing OfficeQA with \textsc{APEX-Agents} already covers static-corpus find, read, and answer work and cross-format deliverable rubrics in a single harness, without a third leaderboard protocol.

\paragraph{Why OfficeQA Pro and \textsc{APEX-Agents}.}
Together they are the closest public pair that satisfies all inclusion criteria while covering complementary regimes: OfficeQA for read-heavy numerical QA over a fixed bulletin library; \textsc{APEX-Agents} for deliverable-heavy professional folders where agents must search an index \emph{and} execute on native files.
Neither is a complete map of knowledge work. GUI suites, parsing leaderboards, and coding benchmarks remain important, but OfficeQA and \textsc{APEX-Agents} are the best available instruments for causal read-axis and review-axis ablations on a single workspace-state stack.

\subsection{Early Design Probes}
\label{app:negative-design}

While building the read surface, we tried simpler layouts on OfficeQA trajectories before the three-way artifact-view ablations in Section~\ref{sec:setup-ablation}.
Three failures were the most instructive.

\paragraph{Parsed companions as the read surface.}
Early builds allowed \texttt{view} on \texttt{*\_parsed/page\_N.md} after \texttt{vector\_search} hits.
Agents often read one indexed page and answered as if they had seen the full bulletin.
We now treat the parsed cache as a \textbf{search layer} only: \texttt{vector\_search} and \texttt{grep} query parsed chunks; \texttt{view} must use the source artifact path and optional \texttt{pages=[...]} (and rejects \texttt{*\_parsed/} paths).

\paragraph{\texttt{grep} on \texttt{*.pdf} without the parsed cache.}
Without a parsed index, \texttt{grep} does not match inside PDF bytes; trajectories show repeated failed \texttt{grep} calls and compensatory \texttt{bash}/\texttt{pdftotext} loops in the \emph{artifact-only} arm.
Keyword search runs over the parsed cache; reads and visual checks use hydrated originals under \texttt{/root/workspace/}.

\paragraph{Static release parse without hash sync.}
OfficeQA's published Databricks \texttt{.txt} export is a strong baseline but ships no live PDFs, no post-edit re-index, and no native execution on the eval workspace. It confounds parser quality with keeping search and sandbox bytes on one tree (Appendix Table~\ref{tab:officeqa-ablation-efficiency}).
Headline ablations use live parsing with a hash-keyed parsed cache; the static export is reported only as a leaderboard-aligned comparison.

\section{Extended Ablation Analysis}
\label{app:extended-analysis}

The full three-arm read-axis ablation and paired-discordance checks appear in Section~\ref{sec:ablation}. This appendix keeps supporting diagnostics: efficiency accounting, the OfficeQA parser-sensitivity check, qualitative coding detail, and residual slices that would crowd the main Results section.

\subsection{Supplementary Efficiency and Parser Checks}
\label{app:residual-design}

The OfficeQA efficiency table adds the released Databricks \texttt{.txt} export as a parser-sensitivity check, while the APEX efficiency table reports only latency, cost, and tool counts to avoid repeating scores from the main Results.

\begin{table}[H]
  \centering
  \appendixtablesetup
  \caption{\textsc{OfficeQA Pro}: artifact-view ablation efficiency. Rows are grouped by model; latency excludes sandbox startup.}
  \label{tab:officeqa-ablation-efficiency}
  \begin{tabular}{llrrrr}
    \toprule
    Model & Configuration & Accuracy & Latency (min) & Cost (\$) & Tools \\
    \midrule
    GPT-5.4 & Artifact-only & \ci{52.6}{8.7} & 8.1 & 0.85 & 34.7 \\
    GPT-5.4 & Parsed-only (Reducto) & \ci{58.6}{8.3} & 17.3 & 0.76 & 34.3 \\
    GPT-5.4 & Parsed-only (Databricks) & \ci{60.2}{8.3} & 7.9 & 0.94 & 30.6 \\
    GPT-5.4 & \textbf{Dual} & \tblbest{\ci{64.7}{8.3}} & 6.1 & 0.84 & 30.4 \\
    \midrule
    Gemini 3 Flash & Artifact-only & \ci{47.7}{8.7} & 16.9 & 0.53 & 41.7 \\
    Gemini 3 Flash & Parsed-only (Reducto) & \ci{53.8}{8.5} & 18.8 & 0.63 & 50.0 \\
    Gemini 3 Flash & Parsed-only (Databricks) & \ci{52.6}{8.7} & 11.6 & 0.71 & 54.5 \\
    Gemini 3 Flash & \textbf{Dual} & \tblbest{\ci{57.8}{8.6}} & 15.3 & 0.48 & 39.8 \\
    \midrule
    Gemini 3.1 Pro & Artifact-only & \ci{55.6}{8.3} & 8.6 & 0.71 & 33.6 \\
    Gemini 3.1 Pro & Parsed-only (Reducto) & \ci{54.9}{8.3} & 18.7 & 0.76 & 33.9 \\
    Gemini 3.1 Pro & Parsed-only (Databricks) & \ci{52.6}{8.7} & 7.3 & 0.78 & 26.7 \\
    Gemini 3.1 Pro & \textbf{Dual} & \tblbest{\ci{63.9}{7.9}} & 10.3 & 0.64 & 25.9 \\
    \bottomrule
  \end{tabular}
\end{table}

\begin{table}[H]
  \centering
  \appendixtablesetup
\caption{\textsc{APEX-Agents}: artifact-view efficiency diagnostics (latency, agent-run cost, and tool counts per arm). Rubric scores are not repeated here; full-pool scores are in Table~\ref{tab:complete-system-results} and paired ablation scores in Table~\ref{tab:dual-artifact-ablation}.}
  \label{tab:apex-sw-efficiency}
  \begin{tabular}{llrrr}
    \toprule
    Model & Arm & Latency (min) & Cost (\$) & Tools \\
    \midrule
    GPT-5.4 & \textbf{Dual} & 5.0 & 1.41 & 28.1 \\
    \midrule
    GPT-5.4 Mini & Artifact-only & 8.8 & 0.33 & 44.0 \\
    GPT-5.4 Mini & Parsed-only & 10.3 & 0.38 & 45.5 \\
    GPT-5.4 Mini & \textbf{Dual} & 7.7 & 0.37 & 42.2 \\
    \midrule
    GPT-5.4 Nano & Artifact-only & 11.7 & 0.13 & 44.3 \\
    GPT-5.4 Nano & Parsed-only & 10.6 & 0.14 & 36.1 \\
    GPT-5.4 Nano & \textbf{Dual} & 11.5 & 0.14 & 39.6 \\
    \midrule
    Gemini 3 Flash & Artifact-only & 5.3 & 0.22 & 31.2 \\
    Gemini 3 Flash & Parsed-only & 5.4 & 0.29 & 23.4 \\
    Gemini 3 Flash & \textbf{Dual} & 3.2 & 0.42 & 28.8 \\
    \midrule
    Gemini 3.1 Pro & \textbf{Dual} & 3.9 & 0.39 & 25.3 \\
    \midrule
    Kimi K2.6 & Artifact-only & 5.5 & 0.39 & 30.5 \\
    Kimi K2.6 & \textbf{Dual} & 5.8 & 0.44 & 32.3 \\
    \bottomrule
  \end{tabular}
\end{table}

\subsection{Qualitative Trajectory Review}
\label{app:qualitative-trajectories}

We code chat logs and grader notes into trajectory labels: which files were opened, which cells or sections were read, and what appeared in the final answer or deliverable.
OfficeQA and \textsc{APEX-Agents} use different label sets matched to each benchmark's trace shape (golden-source pipeline vs.\ workbook/memo deliverables).
Section~\ref{sec:mechanisms} presents the HarFeast trace (Figure~\ref{fig:harfeast-mechanism}); this appendix adds aggregate mechanism counts and an all-arm-zero domain slice rather than reproducing every trajectory checkpoint.

\paragraph{\textsc{APEX-Agents}: Why Dual Scored Higher.}
Among $452$ evaluated tasks, $72$ have task-level best dual above at least one restricted-view arm ($27$ strictly above both).
Table~\ref{tab:apex-dual-mechanisms} codes chat logs on dual-win \emph{model-task instances} (one row per mechanism model and task where dual scores at least as high as both restricted-view arms and strictly beats at least one): $356$ such instances ($253$ unique tasks) vs.\ $147$ strict instances that beat \emph{both} restricted-view arms ($130$ unique tasks).
The instance counts exceed the task-level totals because the same task can dual-win under more than one model.
Percentages in the table are for each instance slice, not across the full task pool.
Mechanism labels are rubric-grounded workspace tags, not OfficeQA golden-source labels.

\begin{table}[H]
  \caption{\textsc{APEX-Agents}: documented mechanisms on dual-win instances. Percentages are within each slice.}
  \label{tab:apex-dual-mechanisms}
  \centering
  \appendixtablesetup
  \begin{tabularx}{0.88\linewidth}{@{}Xr@{}}
    \toprule
    \textbf{Mechanism} & \textbf{Count (\% of slice)} \\
    \midrule
    \multicolumn{2}{@{}l}{\textit{Strict: dual $>$ parsed-only and dual $>$ artifact-only ($n{=}147$ instances)}} \\
    Workspace file discovery (losing arm missed source) & 16/147 (11\%) \\
    Within-file navigation (sheet/section/clause) & 35/147 (24\%) \\
    Table or cell extraction from source & 25/147 (17\%) \\
    Layout or visual read (PDF/deck structure) & 14/147 (10\%) \\
    Model propagation or calculation chain & 28/147 (19\%) \\
    Spreadsheet deliverable execution & 15/147 (10\%) \\
    Document deliverable execution & 3/147 (2\%) \\
    Slide deliverable execution & 1/147 (1\%) \\
    Marginal rubric gain; mechanism unclear & 10/147 (7\%) \\
    \midrule
    \multicolumn{2}{@{}l}{\textit{All dual wins ($n{=}356$ instances): primary route}} \\
    Parsed cache (search, tables, sections) & 148/356 (42\%) \\
    Originals (native files, PDF layout, execution) & 144/356 (40\%) \\
    Both views needed & 33/356 (9\%) \\
    Similar tooling; gap unclear & 31/356 (9\%) \\
    \bottomrule
  \end{tabularx}
\end{table}

\subsection{Residual Failure Slice}
\label{app:errormode}
\label{sec:setup-errormode}

Table~\ref{tab:residual-error-analysis} counts \textsc{APEX-Agents} all-arm-zero tasks by domain ($452$ evaluated tasks).
Section~\ref{sec:residual-failures} interprets this slice; Appendix~\ref{app:qualitative-trajectories} gives compact dual-win traces.

\begin{table}[!htbp]
  \caption{All-arm-zero tasks by \textsc{APEX-Agents} domain ($452$ evaluated tasks). All-arm zero = dual, artifact-only, and parsed-only all score 0 on the same task for at least one completed model triplet.}
  \label{tab:residual-error-analysis}
  \centering
  \resulttablesetup
  \begin{tabular}{lrrr}
    \toprule
    \textbf{Domain} & \textbf{Evaluated} & \textbf{All-arm zero} & \textbf{Rate} \\
    \midrule
    All domains & 452 & 188 & 41.6\% \\
    Investment Banking & 132 & 82 & 62.1\% \\
    Management Consulting & 160 & 55 & 34.4\% \\
    Law & 160 & 51 & 31.9\% \\
    \bottomrule
  \end{tabular}
\end{table}

All-arm-zero counts in Table~\ref{tab:residual-error-analysis} are computed from completed three-arm triplets; they are not trajectory labels.
Section~\ref{sec:residual-failures} interprets these residual failures; compact dual-win traces are in Appendix~\ref{app:qualitative-trajectories}.

\ifdefined\artifactstatepreprint
\section{Benchmark Outlook and Release Notes}
\else
\section{Benchmark Evaluation Outlook}
\fi
\label{app:state-benchmark-protocols}

\ifdefined\artifactstatepreprint
\subsection{Benchmark Evaluation Outlook}
\fi

Section~\ref{sec:discussion} argues that outcome-only scoring limits how sharply workspace-state mechanisms can be validated.
A benchmark for stateful knowledge-work agents should make those mechanisms explicit before scoring final task success.
Following software-engineering benchmarks (Section~\ref{sec:relwork}), knowledge-work agent benchmarks should pair natural-language tasks with inspectable acceptance criteria over files, evidence, and staged edits.
The target extends beyond a better final rubric to include checkpointed scoring of mini-successes (correct source, correct read region, valid staged edit) and task formats that specify which intermediate-state properties are expected and which checks are executable, visual, or human-audited.
The protocol should define:
\begin{itemize}[leftmargin=1.5em,itemsep=2pt,topsep=2pt]
  \item the accepted project files, including paths, content hashes, and which mutations require explicit commit;
  \item the parsed or indexed state available for retrieval, including whether pages, tables, cells, layout regions, comments, tracked changes, formulas, and embedded objects are represented;
  \item the sandbox or execution state available to the agent, including whether generated files, parsed companions, scratch files, and staged edits can be read or modified;
  \item the operation representation for proposed changes, including edit granularity, admissible create/move/delete/replace actions, conflict rules, and source-version requirements;
  \item the verification protocol, including which checks run before commit and whether the agent can inspect their outputs; and
  \item the acceptance or submission rule, including whether a human reviewer, benchmark policy, or automatic grader authorizes the final state.
\end{itemize}
Task construction should also control the distribution of task surfaces, including single-file lookup, multi-file retrieval, spreadsheet edits, slide/document edits, chart or visual reads, and mixed evidence-to-edit workflows, and reserve held-out or private projects when contamination and overfitting are material risks.

Useful labels would include source-version correctness, retrieval-region correctness, evidence-to-edit alignment, staged-operation validity, sandbox-to-journal synchronization correctness, conflict detection, format fidelity, numerical consistency, and verification outcome.
These labels would allow future work to separate reasoning errors from state-management errors.
They would also allow benchmark reports to state which state-transition properties were measured and which were only exercised by the system.

\ifdefined\artifactstatepreprint
  \enlargethispage{10\baselineskip}
\subsection{Reproducibility, Assets, and Responsible-Release Notes}
\else
\section{Reproducibility, Assets, and Responsible-Release Notes}
\fi
\label{app:responsible-release}

\paragraph{Reproducibility Status.}
The experimental design needed to interpret the reported results is described in Section~\ref{sec:setup-benchmarks} (benchmarks, metrics, and ablation design) and Appendix~\ref{app:setup}.
Runs use the original benchmark task sets, graders, and metrics.
The implementation materializes outputs through the reducer before submitting them to the original graders.
Logs retain the operation journal, retrieval traces, diffs, and evidence references for replay and error analysis.

\paragraph{Compute and Cost Accounting.}
The framework does not train new models.
Model inference uses hosted model APIs at the configurations named in Section~\ref{sec:setup-benchmarks}.
Local computation consists of parsing, retrieval, sandboxed Python or shell execution, diff materialization, and grading.
The cost protocol in Appendix~\ref{sec:setup-cost} records per-completion agent-run token charges (excluding background parsing and sandbox host CPU) and wall-clock agent latency at worker concurrency 1.

\paragraph{Existing Assets.}
The experiments use public benchmark assets and graders where released, including \textsc{APEX-Agents} and \textsc{OfficeQA Pro}, cited in the relevant setup and results sections.
This study uses these assets for evaluation without repackaging their data.
The final artifact release should include a license and terms-of-use table for each benchmark, parser, and baseline implementation.

\paragraph{Released Code.}
This paper does not release a new dataset, model, or high-risk pretrained model.
We plan to release a sanitized research implementation of the workspace and evaluation interface via GitHub; it documents workspace materialization, parsed-cache access, tool routing (Table~\ref{tab:agent-tool-routing}), staging, the agent system prompts, and benchmark submission, but was not used for the reported production runs.

\paragraph{Responsible Use.}
\frameworkname{} is intended to make artifact-modifying agents more inspectable by keeping source hashes, tracked file operations, format-aware diffs, and verification traces available during agent work.
These safeguards reduce silent workspace mutation but do not guarantee factual correctness.
High-stakes uses still require qualified human review, especially in legal, financial, clinical, and regulatory settings.

\paragraph{Ethics, Human Subjects, and LLM Usage.}
The work evaluates LLM-based agents on professional-document tasks and does not collect personal data, conduct user studies, or involve human-subject interventions.
Error-mode coding concerns model traces and benchmark artifacts rather than human-subject data.
LLMs are central to the evaluated systems: the models listed in Section~\ref{sec:setup-benchmarks} serve as the agent's policy and reasoning component, while \frameworkname{} provides the workspace, file tracking, and diff surface.

% end NeurIPS checklist \iffalse guard

\end{document}